\documentclass[twoside,11pt]{article}

\usepackage{jmlr2e}
\usepackage{graphicx} 
\usepackage{amssymb}
\usepackage{amsmath}
\usepackage{algorithmicx}
\usepackage[ruled]{algorithm}
\usepackage{algpseudocode}
\usepackage{epstopdf}
\usepackage{subcaption}
\usepackage{hyperref}
\usepackage{caption}
\def\NoNumber#1{{\def\alglinenumber##1{}\State #1}\addtocounter{ALG@line}{-1}}

\algnewcommand\algorithmicinput{\textbf{Assumption:}}
\algnewcommand\Assumption{\item[\algorithmicinput]}

\ShortHeadings{Parallel Algorithm for Structural Discovery in Bayesian Networks}{Chen, Tian, Nikolova, Aluru}
\firstpageno{1}

\begin{document}

\title{A Parallel Algorithm for Exact Bayesian Structure Discovery in Bayesian Networks}

\author{\name Yetian Chen \email yetianc@iastate.edu \\
	   \name Jin Tian \email jtian@iastate.edu \\
       \addr Department of Computer Science\\
       Iowa State University\\
       Ames, IA 50011, USA
       \AND
       \name Olga Nikolova \email olga.nikolova@sagebase.org \\
       \addr Sage Bionetworks\\
       Seattle, WA, USA
	  \AND
	  \name Srinivas Aluru \email aluru@cc.gatech.edu \\
	  \addr School of Computational Science and Engineering \\
		College of Computing, Georgia Institute of Technology \\
	  	Atlanta, GA 30332, USA}

\editor{}
\maketitle

\begin{abstract}
Exact Bayesian structure discovery in Bayesian networks requires exponential time and space.
Using dynamic programming (DP), the fastest known sequential algorithm computes the exact posterior probabilities of structural features in $O(2(d+1)n2^n)$ time and space, 
if the number of nodes (variables) in the Bayesian network is $n$
and the \emph{in-degree} (the number of parents) per node is bounded by a constant $d$.
Here we present a parallel algorithm capable of computing the exact posterior probabilities for all $n(n-1)$ edges with optimal parallel space efficiency and nearly optimal parallel time efficiency.
That is, if $p=2^k$ processors are used, the run-time reduces to $O(5(d+1)n2^{n-k}+k(n-k)^d)$ and the space usage becomes $O(n2^{n-k})$ per processor.
Our algorithm is based the observation that the subproblems in the sequential DP algorithm constitute a $n$-$D$ hypercube.
We take a delicate way to coordinate the computation of correlated DP procedures such that large amount of data exchange is suppressed.
Further, we develop parallel techniques for two variants of the well-known \emph{zeta transform}, which have applications outside the context of Bayesian networks.
We demonstrate the capability of our algorithm on datasets with up to 33 variables and its scalability on up to 2048 processors.
We apply our algorithm to a biological data set for discovering the \emph{yeast} pheromone response pathways. 
\end{abstract}


\section{Introduction}

Bayesian networks (BNs) are probabilistic graphical models that represent 
a set of random variables and their conditional dependencies via a directed
acyclic graph (DAG). Learning the structures of Bayesian networks from data has
been a major concern in many applications of BNs.
In some of the applications,
one aims to find a BN that best explains the observations and then utilizes
this optimal BN for predictions or inferences (we call this \emph{structure learning}). 
In others, we are interested in finding 
the local structural features that are highly probable (we call this \emph{structure discovery}). In causal 
discovery, for example, one aims at the identification of (direct) causal relations among
a set of variables, represented by the edges in the network structure \citep{heckerman1997bayesian}.

Among the approaches to the \emph{structure learning} problem, score-based search method formalizes the problem as an optimization problem 
where a scoring function is used to measure the fitness of a DAG to the observed data, 
then a certain search approach is employed to maximize the score over the space of possible DAGs \citep{cooper1992bayesian,heckerman1997bayesian}. 
While for \emph{structure discovery}, Bayesian method is extensively used. 
In this method, we provide a prior probability distribution $P(G)$
over the space of possible Bayesian networks and compute the posterior distribution
$P(G|D)$ of the network structure $G$ given data $D$. We can then compute the posterior probability
of any structural features by averaging over all possible networks.
Both \emph{structure learning} and \emph{structure discovery} are considered hard since the number of possible
networks is super-exponential, i.e., $O(n!2^{n(n-1)/2})$, with respect to the number of variables $n$. Indeed, it has been showed in \citep{CGH95aistats} that finding an
optimal Bayesian network structure is NP-hard even when the maximum in-degree is bounded by a constant greater than one. 

Recently, a family of DP algorithms have been developed to find the optimal BN in time $O(n2^n)$ and space $O(2^n)$\citep{ott2004finding,singh2005finding,silander2006simple,yuan2011learning,yuan2012improved}.
Likewise, the posterior probability of structural features can be computed by analogous DP techniques. 
For example, the algorithms developed in \citep{koivisto2004exact} and \citep{Koivisto06} can compute the exact marginal
posterior probability of a subnetwork (e.g., an edge)
and the exact posterior probabilities for all $n(n-1)$ potential edges in $O(n2^n)$ time and space,
assuming that the \emph{in-degree}, i.e., the number of parents of each node, is bounded by a constant.
However, these algorithms require a special form of the structural prior (called order modular prior), which deviates from the simplest uniform prior and does not respect Markov equivalence
\citep{friedman2003being}. If adhering to the standard (structure modular) prior, the fastest known algorithm is slower, taking time $O(3^n)$ and space $O(n2^n)$ \citep{tian2009computing}.
Due to their exponential time and space complexities, the largest networks these DP algorithms can solve on a typical desktop computer with a few GBs of memory do not exceed 25 variables.

While both the time and space requirements grow exponentially as $n$ increases, it is the space requirement being the bottleneck in practice. 
Several techniques have been developed to reduce the space usage. In \citep{malone2011memory}, the DP algorithm  
for finding optimal BNs in \citep{singh2005finding} is improved such that only the scores and information for two adjacent layers in the recursive graph 
are kept in memory at once. This manipulation reduces the memory usage to $O({n\choose n/2})$. 
And they showed the implementation of the algorithm solved a problem of 30 variables in about 22 hours using 16 GB memory. 
However, their implementation needs external memory (i.e., hard disk) to store the entire recursive graph. 
This may slow down the algorithm due to the slow access to hard disk. 
Further, the algorithm is not scalable on larger problems as the $O({n\choose n/2})$ space usage still grows very fast as $n$ increases.
Alternatively, \citet{parviainen2009exact} proposed several schemes to trade space against time. If little space is available, a divide-and-conquer scheme recursively splits the problem to subproblems, 
each of which can be solved completely independently. This scheme results in time $2^{2n-s}n^{O(1)}$ in space $2^sn^{O(1)}$ for any $s=n/2, n/4, n/8, ...$, where $s$ is the size of the subproblems.
If moderate amounts of space are available, a pairwise scheme splits the search space by fixing a class of partial orders on the set of variables. 
This manipulation yields run-time $2^n(3/2)^pn^{O(1)}$ in space $2^n(3/4)^pn^{O(1)}$ for any $p=0,1,...,n/2$ where $p$ is a parameter controlling the space-time trade-off.
Although both schemes make it practical to solve larger problems using limited space, they make a huge sacrifice in running time.  

Parallel computing aims to design systems and algorithms that use multiple processing elements simultaneously to solve a problem.
It allows us to overcome the time and space limitations by using supercomputers, which are usually equipped with thousands of processors and several terabytes memory.
If the computation steps in solving a problem are independent, the running time can be significantly reduced by parallelizing the execution of these independent steps on multiple processors. 
Certainly, this acceleration has theoretical upper bound. 
A widely used measure of the acceleration is \emph{speedup}, 
defined as the ratio between the sequential running time (on one processor) and the parallel running time on $p$ processors. 
Then in theory, $speedup \leq p$. That is, one can't achieve more than $p$ times faster if $p$ processors are used.
The $speedup$ will often be less than $p$ as the parallel algorithm is bound to have some overhead in coordinating the actions of processors.
Another measure of the performance of a parallel algorithm is \emph{efficiency}, 
defined as the ratio between the sequential running time and the product of the number of processors used and the parallel running time.
\emph{Efficiency} measures how well the processors are utilized by the algorithm. 
Similarly, \emph{efficiency} $\leq 1$. A parallel algorithm is said to be efficient if it involves the same order of work as performed by the best sequential algorithm.
Most modern supercomputers implement a parallel model called the \emph {distributed memory model}\footnote{Another popular parallel model is the \emph{shared memory model}, 
where a memory space is shared by all processors. This type of systems is typically very expensive and not scalable in terms of the memory size and the number of processors.}, 
where many processors are linked through high-speed
connections and each processor has local memory directly attached to it. 
This type of supercomputers is scalable in terms of both the memory space and the number of processors.
Thus, current research in parallel computing mainly use the \emph{distributed memory model} for designing parallel algorithms.


Several parallel algorithms have already been developed for solving the \emph{structure learning} problem. 
First, as mentioned by the authors, the pairwise scheme proposed in \citep{parviainen2010bayesian} allows easy parallelization on up to $2^p$ processors for any $p=0,1,...,n/2$. Each of the processors solves a subproblem independently in time $2^n(3/4)^pn^{O(1)}$ in space $2^n(3/4)^pn^{O(1)}$.
Compared to the sequential algorithm that runs in time and space of $2^nn^{O(1)}$, the parallel efficiency is $(2/3)^p$, which is suboptimal. 
Further, they only implemented the pairwise scheme to compute the subproblems. Thus, although their results suggest the implementation is feasible up to around 31 variables, their estimate ignores the parallelization overhead that generally becomes problematic in parallelization.
Later, \citet{tamada2011parallel} presented a parallel algorithm that splits the search space so that the required communication between subproblems is minimal. 
The overall time and space complexity is $O(n^{\sigma+1}2^n)$, where $\sigma=0,1..., >0$ controls the communication-space trade-off. 
This algorithm, as mentioned, has slightly greater space
and time complexities than the algorithm in \citep{parviainen2009exact} because of redundant calculations of DP steps. 
Their implementation of the algorithm was able to solve 32-node network in about 5 days 14 hours using 256 processors with 3.3 GB memory per processor. 
However, it did not scale well on more than 512 processors as the parallel efficiency decreased significantly from 0.74 on 256 processors to 0.39 on 1024 processors. 
\citet{nikolova2009parallel,nikolova2013parallel} described a novel parallel algorithm that realizes direct parallelization of the sequential DP algorithm in \citet{ott2004finding} with optimal parallel efficiency. This algorithm is based on the observation that the subproblems constitute a lattice equivalent to an $n$-dimensional ($n$-$D$) hypercube, 
which has been proved to be a very powerful interconnection
network topology used by most of modern parallel computer systems \citep{dally2004principles,ananth2003introduction,loh2005exchanged}.
In the lattice formed by the DP subproblems, data exchange only happens between two adjacent nodes.
In a hypercube interconnection network, the neighbors communicate with each other much more efficiently than other pairs of nodes.
By noting this, the parallel algorithm takes a direct mapping of the DP steps to the nodes of a hypercube, thus is communication-efficient.
Further, this hypercube algorithm does not calculate redundant steps or scores.  
These two features render the implementation of the algorithm scalable on up to 2048 processors \citep{nikolova2013parallel}.
Using 1024 processors with 512 MB memory per processor, they solved a problem with 30 variables in 1.5 hours.  

In contrast, using parallel computing to speed and scale up \emph{structure discovery} has not been studied so extensively.
To our knowledge, there are no parallel algorithms developed for computing the exact posterior probability of structural features.
Although \citet{parviainen2010bayesian} extended the parallelizable partial-order scheme to the \emph{structure discovery} problem,
they did not offer any explicit way to parallelize it. Although the DP techniques for these two problems are analogous, they differ in some significant places.
These differences prohibit the direct adaption of the existing parallel algorithms for \emph{structure learning} to \emph{structure discovery}.
First, the DP algorithm for optimal BN learning involves only one DP procedure. All relevant scores for a certain subproblem are computed in one DP step, 
therefore can be computed on one processor. Thus, the mapping of subproblems to processors is very straightforward.
However, the DP algorithm for computing the posterior probability of structural features involves several separate DP procedures, responsible for computing different scores.
These DP procedures, though can be performed separately, rely on the completion of one another.
Thus, it is a challenge to effectively coordinate the computations of these DP procedures in a parallel setup. Failure to do this may greatly harm the parallel efficiency. 
Second, the DP algorithm for computing the posterior probability of structural features involves two critical subtasks, 
each of which calls for a fast computation of a zeta transform variant.
These two zeta transform variants require efficient parallel processing. 

To fill up the gap, in this paper we develop a parallel algorithm to compute the exact posterior probability of substructures (e.g., edges) in Bayesian networks.
Our algorithm realizes direct parallelization of the DP algorithm in \citep{Koivisto06} with nearly perfect load-balancing and optimal parallel time and space efficiency, i.e., the time and space complexity per processor are $O(n2^{n-k})$ respectively, for $p$ number of processors, where $k=log(p)$. 
Our parallel algorithm is an extension of \citet{nikolova2009parallel}'s hypercube algorithm 
to the \emph{structure discovery} problem. However, because of the difficulties discussed previously, our work goes beyond that by a significant margin. First, we adopt a delicate way to map the calculation of various scores to the processors such that large amount of data exchange between non-neighboring processors is avoided during the transition among the separate DP procedures. This manipulation significantly reduces the time spent in communication. Second, we develop novel parallel algorithms for two fast zeta transform variants. As zeta transforms are fundamental objects in several several combinatorial problems such as graph coloring \citep{koivisto20062} and Steiner tree \citep{nederlof2009fast}
and combinatorial tools like the fast subset convolution \citep{bjorklund2007fourier}, 
the parallel algorithms developed here would also benefit the researches outside the context of Bayesian networks. 

The rest of the paper is organized as follows. In Section 2, we present some preliminaries of exact structure discovery in BNs
and briefly review \citet{Koivisto06}'s DP algorithm, upon which our parallel algorithm is based. In Section 3, we present our parallel algorithm for 
computing the posterior probability of structural features and conduct a theoretical analysis on its run-time and space complexity. 
In Section 4, we empirically demonstrate the capability of
our algorithm on a Dell PowerEdge C8220 supercomputer. Discussions and conclusions are presented in Section 5.

\section{Exact Bayesian Structure Discovery in Bayesian Networks} 

Formally, a Bayesian network is a DAG that encodes a joint probability distribution over a vector of random variables $\mathbf{x} = (x_1,...,x_n)$
with each node of the graph representing a variable in $\mathbf{x}$. For convenience we will typically work on the index set $V = {\{1,...,n\}}$
and represent a variable $x_i$ by its index $i$. The DAG is represented as a vector $G=(G_1,...,G_n)$ where each $G_i$ is a subset
of the index set $V-\{i\}$ and specifies the parents of $i$ in the graph.

Given a set of observations $D$, the joint distribution $P(G,D)$ is computed by

\begin{equation}{P(G,D) = P(G)P(D|G)},\end{equation}
where $P(G)$ specifies the structure prior and $P(D|G)$ is the likelihood of the data.

\subsection{Computing Posteriors of Structural Features}  

In this section, we review the DP algorithm in \citep{koivisto2004exact} for computing the posteriors of structural features.
A structural feature, e.g., an edge, is conveniently represented by an indicator function $f$ such that $f(G)$ is 1~if the feature is present in $G$ and 0~otherwise.
In Bayesian approach, we are interested in computing the posterior probability $P(f|D)$ of the feature, which can be obtained by computing the joint probability $P(f,D)$ as

\begin{equation}{P(f,D) = \sum\limits_{G}{f(G)P(G,D)}}.\end{equation}

Instead of directly summing over the super-exponential DAG space, \citet{friedman2003being} proposed to work on the order space, which was
demonstrated more efficient and convenient. Formally, an order $\prec$ is a linear order $(L_1,...,L_n)$ on the index set $V$,
where $L_i$ specifies the predecessors of $i$ in the order, i.e., $L_i=\{j:j{\prec}i\}$. We say that a DAG $G=(G_1,...,G_n)$ is consistent
with an order $\prec$ if $G_i{\subseteq}L_i$ for all $i$. By introducing the random variable $\prec$, $P(f,D)$ can be computed alternatively by

\begin{equation}{P(f,D) = \sum_{\prec}{P({\prec})P(f,D|{\prec})}}.\label{eq:jointpfd}\end{equation}

Assume an \emph{order modular prior} defined as follows: if $G$ is consistent with $\prec$, then

\begin{equation}{P({\prec},G) = \prod\limits_{i=1}^nq_i(L_i){\rho}_i(G_i)},\end{equation}
where each $q_i$ and $\rho_i$ is some function from the subsets of $V-\{i\}$ to the nonnegative reals. 
We will also make the standard assumptions on global and local parameter independence, and parameter modularity \citep{cooper1992bayesian}.
Further, in this paper, we consider only modular features, i.e, $f(G) = \prod_{i=1}^nf_i(G_i)$, where each $f_i(G_i)$ is an indicator 
function with values either 0 or 1. For example, an edge $u \rightarrow v$ can be represented by setting $f_v(G_v)=1$ if and only if $u \in G_v$,
and setting $f_i(G_i)=1$ for all $i \neq v$. 
In addition, we assume the number of parents of each node is bounded by a constant $d$. With these assumptions, \citet{koivisto2004exact} showed that Eq.~(\ref{eq:jointpfd}) can be factorized as

\begin{equation}
P(f,D) = 
\sum\limits_{\prec}{\prod\limits_{i=1}^nq_i(L_i)\sum\limits_{G_i:~G_i \subseteq L_i \text{~and~} |G_i|\leq d}\rho_i(G_i)p(x_i|x_{G_i},G_i)f_i(G_i)},\label{eq:factorize}
\end{equation}
where $p(x_i|x_{G_i},G_i)$ is the local marginal likelihood for variable $i$, measuring the local goodness of $G_i$ as the parents of $i$.
For convenience, for each family $(i, G_i)$, $i\in V$, $G_i\subseteq V-\{i\}$, we define

\begin{equation}B_i(G_i)\equiv\rho_i(G_i)p(x_i|x_{G_i},G_i)f_i(G_i).\end{equation}

Note that if we assume the bounded in-degree $d$, we only need to compute $B_i(G_i)$ for $G_i\subseteq V-\{i\}$ with $|G_i|\leq d$.
Further, for all $i\in V$, $S\subseteq V-\{i\}$, define

\begin{equation}
A_i(S)\equiv q_i(S)\sum\limits_{G_i:~G_i \subseteq S\text{~and~}|G_i|\leq d}B_i(G_i).
\label{eq:A}
\end{equation}

The sum on the right-hand side of Eq.~(\ref{eq:A}) is known as a variant of the \emph{zeta transform} of $B_i$, evaluated at $S$.
Now Eq.~(\ref{eq:factorize}) can be neatly written as

\begin{equation}{P(f,D) = \sum\limits_{\prec}{\prod\limits_{i=1}^nA_i(L_i)}}.\end{equation}

\citet{koivisto2004exact} showed that $P(f,D)$ can be computed by defining a recursive function $F$ on all $S\subseteq V$,

\begin{equation}{F(S)\equiv \sum\limits_{i \in S}A_i(S-\{i\})F(S-\{i\})},\end{equation}
with the base case $F(\emptyset)\equiv 1$. Then $P(f,D)=F(V)$.

With this definition, $P(f,D)=F(V)$ can be computed efficiently with dynamic programming. Then the posterior probability of the feature $f$ is 
obtained by $P(f|D)=P(f,D)/P(D)$, where $P(D)$ can be computed like $P(f,D)$ by simply setting all features $f_i(G_i) = 1$, i.e.
$P(f=1,D)=P(D)$.

Computing $B_i(G_i)$ scores for all $i\in V$, $|G_i|\leq d$ takes $O(n^{d+1})$ time.\footnote{ We assume the computation of $p(x_i|x_{G_i},G_i)$ takes $O(1)$ time here. However, 
it is usually proportional to the sample size $m$.} For any $i\in V$, 
$A_i$ scores can be computed in $O(d2^n)$ time with a technique called 
the {\emph{fast truncated upward zeta transform}\footnote{It is called M\"{o}bius transform in \citep{koivisto2004exact,Koivisto06}, but \emph{zeta transform} is actually the correct term.}
\citep{koivisto2004exact}. The recursive computation of
$F(V)$ takes $O(\sum_{i=0}^ni\cdot{n\choose i}) = O(n2^n)$ time. The total time for computing one feature (i.e., an edge) is therefore $O(n^{d+1}+nd2^n+n2^n)=O((d+1)n2^n)$.

\subsection{Computing Posterior Probabilities for All Edges}   

If the application is to compute the posteriors for all $n(n-1)$ edges, we can run above algorithm separately for each edge. Then the time for computing all $n(n-1)$ edges is $O((d+1)n^32^n)$. Since the computations for different edges involve a large proportion of overlapping elements, a forward-backward algorithm was provided in \citep{Koivisto06} to reduce the
time to $O(2(d+1)n2^n)$.
For all $S\subseteq V$, define a ``backward function" recursively as





\begin{equation}{R(S)=\sum\limits_{i \in S}A_i(V-S)R(S-\{i\})},\end{equation}
with the base case $R(\emptyset)=1$.
Then for any fixed node $v \in V$ (the endpoint of an edge) and $u \in V-\{v\}$, the joint distribution $P(u\rightarrow v,D)$ can be computed by

\begin{equation}{P(u\rightarrow v,D)=\sum\limits_{G_v:~u\in G_v \subseteq V-\{v\} \text{~and~} |G_v|\leq d}B_v(G_v)\Gamma_v(G_v)}, \label{ed:jointedge}\end{equation}
where for all $v\in V$, $G_v\subseteq V-\{v\}$
\begin{equation}{\Gamma_v(G_v)\equiv\sum\limits_{S:G_v \subseteq S\subseteq V-\{v\}}q_v(S)F(S)R(V-\{v\}-S)}.\label{eq:gamma}\end{equation}

The sum on the right-hand side of Eq.~(\ref{eq:gamma}) is another variant of the \emph{zeta transform}.
Provided that $B_i$, $A_i$, $F$, $R$ are precomputed with respect to $f\equiv 1$, for any endpoint node $v$, $\Gamma_v(G_v)$ can be computed in $O(d2^n)$ time for all $G_v\subseteq V-\{v\}$,
$|G_v|\leq d$ with a technique called \emph{fast downward zeta transform} \citep{Koivisto06}. 
To evaluate Eq.~(\ref{ed:jointedge}) for a different $u\in V-\{v\}$, we only need to recompute the function $B_v$ by changing only the function $f_v$. 
Thus, evaluating Eq.~(\ref{ed:jointedge}) takes $O(n^d)$ time. 

We then arrived at the following algorithm for computing the posteriors for all $n(n-1)$ edges. Let the functions $B_i$, $A_i$, $\Gamma_i$, $F$ and $R$ be defined
with respect to the trivial feature $f\equiv 1$ ($f_i(G_i)=1$ for all $i\in\{1,...,n\}$ and $G_i\subseteq V-\{i\}$).

\begin{algorithm}[H]
\caption{Compute posterior probabilities for all $n(n-1)$ edges by DP \citep{Koivisto06}}
\label{algorithm1}
\begin{algorithmic}[1]
\State \textbf{for} all $i\in V$ and $S\subseteq V-\{i\}$ with $|S|\leq d$: compute $B_i(S)$.
\State \textbf{for} all $i\in V$ and $S\subseteq V-\{i\}$: compute $A_i(S)$.
\State \textbf{for} all $S\subseteq V$: compute $F(S)$ recursively.
\State \textbf{for} all $S\subseteq V$: compute $R(S)$ recursively.
\For{all $v\in V$}
\State \textbf{for} all $G_v\subseteq V-\{v\}$ with $|G_v|\leq d$: compute $\Gamma_v(G_v)$.
\For{all $u\in V-\{v\}$} 
\State Compute $P(u\rightarrow v,D)=\sum\limits_{G_v:~u\in G_v \subseteq V-\{v\}\text{~and~}|G_v|\leq d}B_v(G_v)\Gamma_v(G_v)$
\State Evaluate $P(u\rightarrow v|D)=P(u\rightarrow v,D)/F(V)$.
\EndFor
\EndFor
\end{algorithmic}
\end{algorithm}

Adding up the time for all steps, the total computation time for evaluating all $n(n-1)$ edges is $O(n^{d+1}+dn2^n+n2^n+n2^n+n^{d+2}+dn2^n)=O(2(d+1)n2^n)$.

\section{Parallel Algorithm} 

We use the Algorithm~\ref{algorithm1} presented in Section 2 as a base for parallelization.
The computation of $A$ functions corresponds to a variant of \emph{zeta transform} which are computationally intensive. 
There is no known parallel algorithm for \emph{zeta transform} \citep{rota1964foundations}. Here we design a novel parallel algorithm for it.\footnote{The $B$ functions required for computing $A$ functions are computed inside the algorithm for computing $A$ functions.}
The recursive computations of functions $F$ and $R$ over node sets $S\subseteq V$ are analogous to DP techniques in the algorithm for finding optimal BN in \citep{ott2004finding}. 
Thus, it is possible to use the parallel techniques developed for the latter problem.
In our algorithm, we adapt \citet{nikolova2009parallel}'s hypercube algorithm and use it as sub-routines to compute functions $F$ and $R$.
However, some difficulties prohibit the direct adaption. 
The computation in Algorithm~\ref{algorithm1} consists of several consecutive procedures, each of which is responsible for computing a particular function.
The computations of these functions depend on one another. For example, computing $F$ and $R$ need $A$'s being available.
Note that all functions are evaluated over $2^n$ of subsets $S\subseteq V$. In the hypercube algorithm, these subsets are computed in different processors, 
thus stored locally. Generally, processors need exchange their scores in order to compute a new function. 
Thus, it is challenging to coordinate the computations of these functions to reduce the number of messages sent between the processors, 
particularly between those non-neighboring processors. In our algorithm, we adopt a delicate way to achieve this.
The computations of $\Gamma$ functions correspond to another variant of \emph{zeta transform} for which we again design a novel parallel algorithm.
Finally, we integrate these techniques into a parallel algorithm capable of computing the posteriors $P(u\rightarrow v|D)$ for all $n(n-1)$ edges
with nearly perfect load-balancing and optimal parallel efficiency.

To facilitate presentation, in Section 3.1, we first describe an ideal case, where $2^n$ processors are available.
In this case, we can directly map the $2^n$ of subsets $S\subseteq V$ to an $n$-dimensional hypercube computing cluster. 
In Section 3.2, we then generalize the mapping to a $k$-dimensional hypercube with $k<n$.
 
\subsection{$n$-$D$ Hypercube Algorithm}
In Section 3.1.1, we first describe the parallel algorithms for computing functions $F$ and $R$ as they explain why we base our parallel algorithm on the hypercube model. We postpone the discussion of computing $B$, $A$ scores to Section 3.1.2.

\subsubsection{Computing $F(S)$ and $R(S)$}

The DP algorithm for computing functions $F$ can be visualized as operating on the lattice $\mathcal{L}$ 
formed by the partial order ``set inclusion" on the power set of $V$ (see Figure~\ref{fig1}).
The lattice $\mathcal{L}$ is a directed graph $(V',E')$, where $V'=2^V$ and $(T,S)\in E'$ if $T\subset S$ and $|S|=|T|+1$. 
The lattice is naturally partitioned into levels, where level $l$ ($0 \le l\le n$) contains all subsets of size $l$. 
A node $S$ at level $l$ has $l$ incoming edges from nodes $S-\{i\}$ for each $i\in S$, and $n-l$ outgoing edges to nodes $S\cup\{j\}$ for each $j\notin S$. 
By Eq. (9), node $S$ receives $A_i(S-\{i\})\cdot F(S-\{i\})$ from each of its incoming edges and computes $F(S)$ by summing over $l$ such scores. 
Assuming $A_j(S)$ for all $j\notin S$ are precomputed and available at node $S$, 
node $S$ will compute $A_j(S)\cdot F(S)$ for all $j\notin S$ then send the scores to corresponding nodes. For example, $A_j(S)\cdot F(S)$ is sent
to node $S\cup \{j\}$ so that it can be used for computing $F(S\cup \{j\})$. Each level in the lattice can be computed concurrently, with data flowing from one level to the next.


If each node in $\mathcal{L}$ is mapped to a processor in a computer cluster, 
the undirected version of $\mathcal{L}$ is equivalent to an $n$-dimensional ($n$-$D$) hypercube, 
a network topology used by most of modern parallel computer systems \citep{dally2004principles,ananth2003introduction,loh2005exchanged}.
We encode a subset $S$ by an $n$-bit string $\omega$, where $\omega[i]=1$ if variable $i\in S$ and $\omega[i]=0$ otherwise.
Accordingly, we can use $\omega$ to denote the \emph{id} of the processor that the subset $S$ is mapped to. 
As lattice edges connect pairs of nodes whose $n$-bit string differ by one element, they naturally correspond to hypercube edges (Figure~\ref{fig1}). 
This suggests an obvious parallelization on an $n$-$D$ hypercube.

The $n$-$D$ hypercube algorithm runs in $n+1$ steps. Let $\mu(\omega)$ denote the number of 1's in $\omega$. 
Each processor is active in only one time step
-- processor $\omega$ is active in time step $\mu(\omega)$. 
It receives one $A_i(S-\{i\})\cdot F(S-\{i\})$ value from each of $\mu(\omega)$ neighbors obtained by inverting one of its 1 bits to 0. 
It then computes its $F(S)$ function, computes $A_j(S)\cdot F(S)$ for all $j\notin S$ and sends them to its $n-\mu(\omega)$ neighbors obtained by inverting one of its 0 bits to 1. 
The run-time of step $l$ is $O(l+n-l)=O(n)$. The parallel run-time for computing all $F$ scores is $O(n^2)$ in total.

We can parallelize the computation of function $R$ in the same manner. However, we have assumed $A_j(S)$ for all $j\notin S$ are available only at node $S$.
To compute $R(S)$, node $S$ need receive $A_i(V-S)\cdot R(S-\{i\})$ from its neighbors. 
However, $A_i(V-S)$ are available at neither node $S-\{i\}$ nor node $S$, but node $V-S$. 
Further, it is not a good idea either to have $F(S)$ and $R(S)$ at the same processor 
as each term of the summation in the computation of $\Gamma$ scores requires different $F$ and $R$ (see Eq.~(12)). 
To reduce message passing, we take a completely different mapping for computing $R$. The new mapping is illustrated in Figure~\ref{fig2}.
Note that $R(S)$ is computed at the processor where $F(V-S)$ is computed and all $A_i(V-S)$ are available. 
Processor $\omega$ receives one $R(S-\{i\})$ from each of its $n-\mu(\omega)$ neighbors obtained by inverting one of its 0 bits to 1. 
It then computes $R(S)$ by Eq.~(10) and sends it to all its $\mu(\omega)$ neighbors obtained by inverting one of its 1 bits to 0. 
The processors in the hypercube operate in a bottom-up manner, e.g., starting from processor $111$ and ending at processor $000$. 
Similarly, the parallel run-time is $O(n^2)$. 

\begin{figure*}
	\begin{minipage}[t]{3in}
		\begin{figure}[H]
			\centering
			\includegraphics[width=60mm]{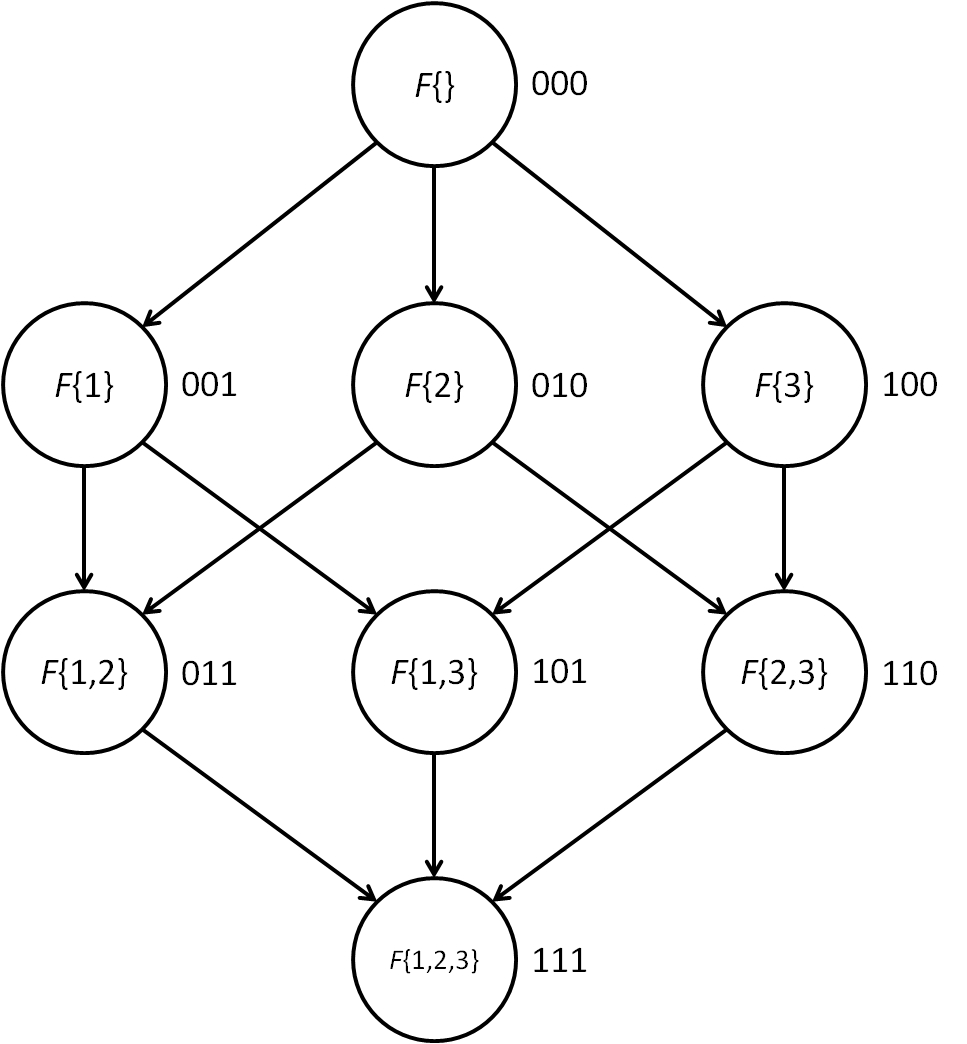}
			\caption{\small A lattice for a domain of size 3. The binary string labels on the right-hand side of each node show the correspondence with a 3-dimensional hypercube. $B_i$, $A_i$ and function $F$ for each subset $S$ are computed at corresponding processor. The arrows show how the data flow between subproblems.}
			\label{fig1}
		\end{figure}
	\end{minipage}
	\hfill
	\begin{minipage}[t]{2.7in}
		\begin{figure}[H]
			\centering
			\includegraphics[width=60mm]{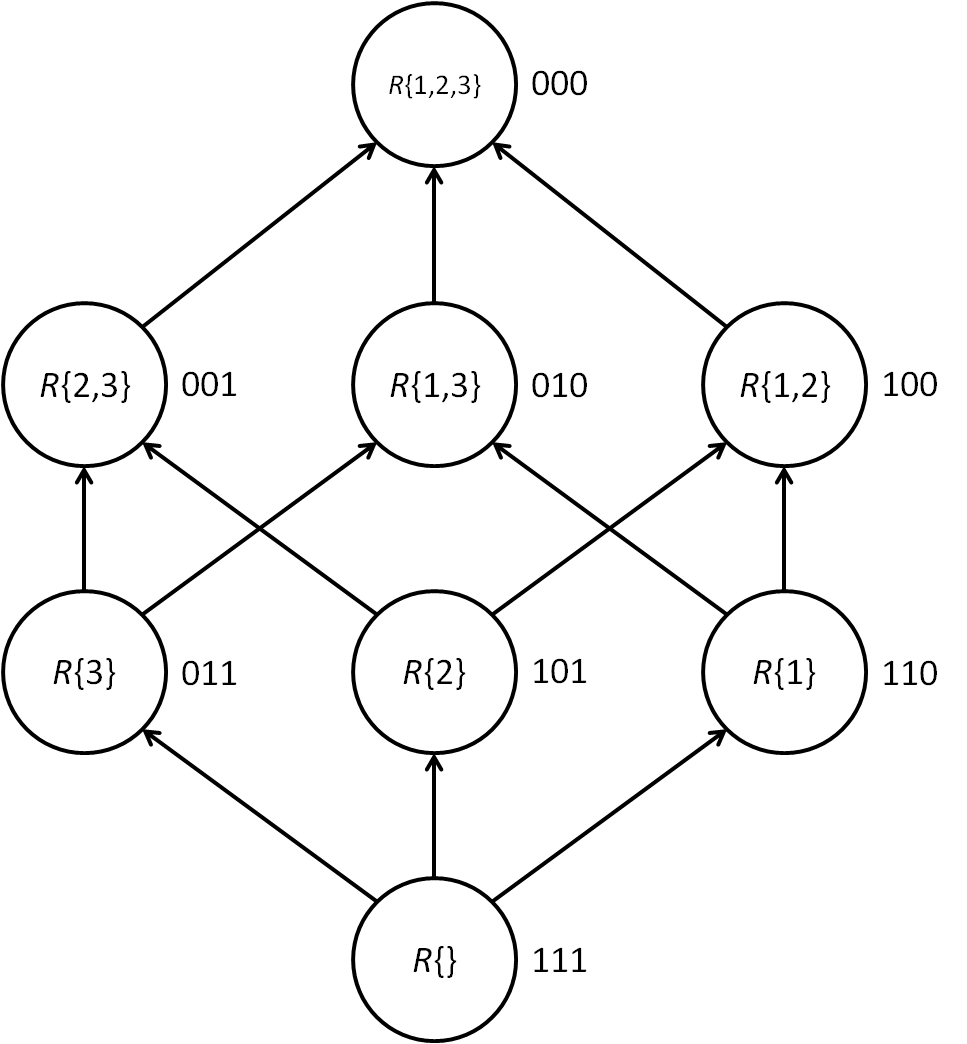}
			\caption{\small Map the computation of function $R(S)$ to the $n$-$D$ hypercube. The binary string label on the right-hand side of each node denote the \emph{id} of the processor.
				The arrows show how the data flow between subproblems.}
			\label{fig2}
		\end{figure}
	\end{minipage}
	\hfill
	\vspace{-1em}
\end{figure*}

\subsubsection{Parallel Fast Zeta Transforms}

In Section 3.1.1, we have assumed $A_i(S)$ for all $i\notin S$ are precomputed at node $S$. 
For any $i\in V$, computing $A_i(S)$ for any subset $S \subseteq V-\{i\}$ requires the summation over all subsets of $S$ with size no more than $d$ (see Eq.~(7)). 
If processors in the hypercube compute their $A_i$ independently, the processor responsible for computing $A_i(V-\{i\})$ for all $i\in V$ takes $O(d2^{n-1})$ time. 
This certainly nullifies our effort of improving time complexity by parallel algorithm.
In this section, we describe parallel algorithms with which all $A_i$ (and $\Gamma_v$) scores can be computed on the $n$-$D$ hypercube cluster in $O(n^2)$ time.

First, we give definitions for two variants of the well-known \emph{zeta transform} \citep{kennes1992computational}. 
Let $V=\{1,...,n\}$. Let $s: 2^V\rightarrow\mathbb{R}$ be a mapping from the subsets of $V$ onto the real numbers. Let $d\leq |V|$ be a positive integer.

\begin{definition}
\citep{koivisto2004exact}:  A function $t: 2^V\rightarrow\mathbb{R}$ is the truncated upward zeta transform of $s$ if
\begin{align*}t(T)=\sum\limits_{S\subseteq T:|S|\leq d}s(S) \text{,~for all~} T\subseteq V.\end{align*}
\end{definition}

\begin{definition}
\citep{Koivisto06}:  A function $t: 2^V\rightarrow\mathbb{R}$ is the truncated downward zeta transform of $s$ if
\begin{align*}t(T)=\sum\limits_{S:T\subseteq S\subseteq V}s(S) \text{,~for all~} T\subseteq V \text{~with~} |T|\leq d.\end{align*}
\end{definition}

It is easy to see that the function $A_i$ for all $i\in V$ can be viewed as a case of the truncated \emph{upward} zeta transform. 
Similarly, the function $\Gamma_v(G_v)$ can be viewed as a case of the truncated \emph{downward} zeta transform.

Two techniques introduced in \citep{koivisto2004exact} and \citep{Koivisto06} are able to realize both transforms in $O(d2^n)$ time, respectively. 
Here, we present the parallel versions of the two algorithms (see Algorithm~\ref{algorithm2} and Algorithm~\ref{algorithm3}) that run on an $n$-$D$ hypercube computer cluster.
The serial versions of the algorithms are given in \citep{koivisto2004exact} and \citep{Koivisto06}, respectively.

\begin{algorithm}[H]\small
	\caption{Parallel Truncated Upward Zeta Transform on $n$-$D$ hypercube}
	\label{algorithm2}
	\begin{algorithmic}[1]
		\Assumption each subset $S\subseteq V$ is encoded by an $n$-bit string $\omega$, where $\omega[i]=1$ if variable $i\in S$ and $\omega[i]=0$ otherwise. Subset $S$ is computed on processor with \emph{id} $\omega$.
		\State On each , $t_0(S)\leftarrow s(S)$ for $|S|\leq d$ and $t_0(S)\leftarrow 0$ otherwise.
		\For{$j\leftarrow 1$ to $n$}
		\For{each processor $\omega$ s.t. $|S\cap\{j+1,...,n\}|\leq d$}
		\State $t_j(S)\leftarrow 0$
		\If{$|S\cap \{j,...,n\}|\leq d$}
		\State $t_j(S)\leftarrow t_{j-1}(S)$
		\EndIf
		\If{$j\in S$}
		\State Retrieve $t_{j-1}(S-\{j\})$ from processor $\omega'=\omega\oplus2^{j-1}$. 
		\State $t_j(S)\leftarrow t_j(S)+t_{j-1}(S-\{j\})$
		\EndIf
		\EndFor
		\EndFor
		\State \textbf{return} $t_n(S)$ on processor $\omega$
	\end{algorithmic}
\end{algorithm}


\begin{algorithm}[H]\small
	\caption{Parallel Truncated Downward Zeta Transform on $n$-$D$ hypercube}
	\label{algorithm3}
	\begin{algorithmic}[1]
		\Assumption each subset $S\subseteq V$ is encoded by an $n$-bit string $\omega$, where $\omega[i]=1$ if variable $i\in S$ and $\omega[i]=0$ otherwise. Subset $S$ is computed on processor with \emph{id} $\omega$.
		\State On each processor, $t_0(S)\leftarrow s(S)$.
		\For{$j\leftarrow 1$ to $n$}
		\For{each processor $\omega$ s.t. $|S\cap\{1,...,j\}|\leq d$}
		\State $t_j(S)\leftarrow t_{j-1}(S)$
		\If{$j\notin S$}
		\State Retrieve $t_{j-1}(S\cup\{j\})$ from processor $\omega'=\omega\oplus2^{j-1}$. 
		\State $t_j(S)\leftarrow t_j(S)+t_{j-1}(S\cup\{j\})$
		\EndIf
		\EndFor
		\EndFor
		\State \textbf{return} $t_n(S)$ on processor $\omega$
	\end{algorithmic}
\end{algorithm}

\begin{figure}
	\centering
	\begin{subfigure}[b]{0.48\linewidth}
		\centering
		\includegraphics[width=0.95\linewidth]{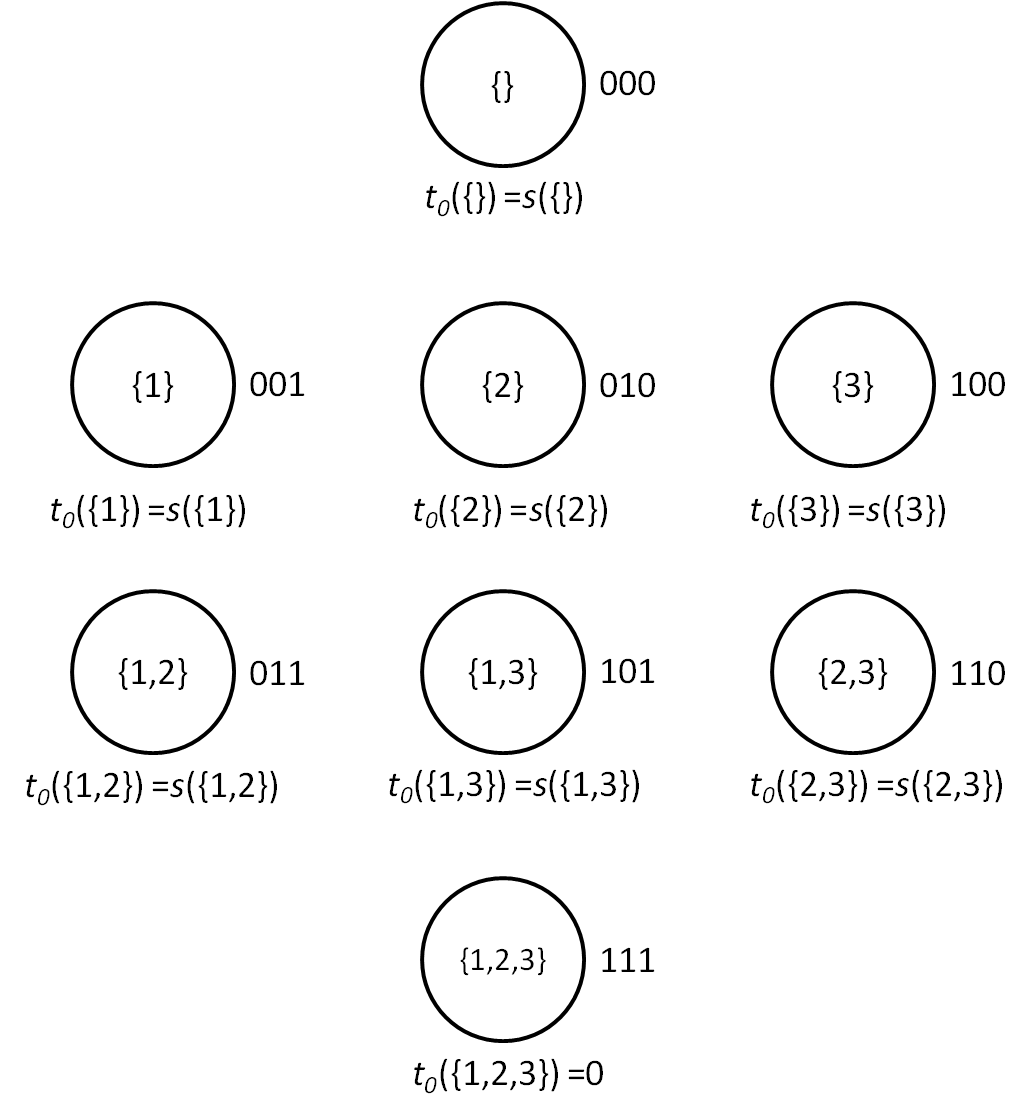}
		\caption{$j=0$}
	\end{subfigure}
	\begin{subfigure}[b]{0.48\linewidth}
		\centering
		\includegraphics[width=0.92\linewidth]{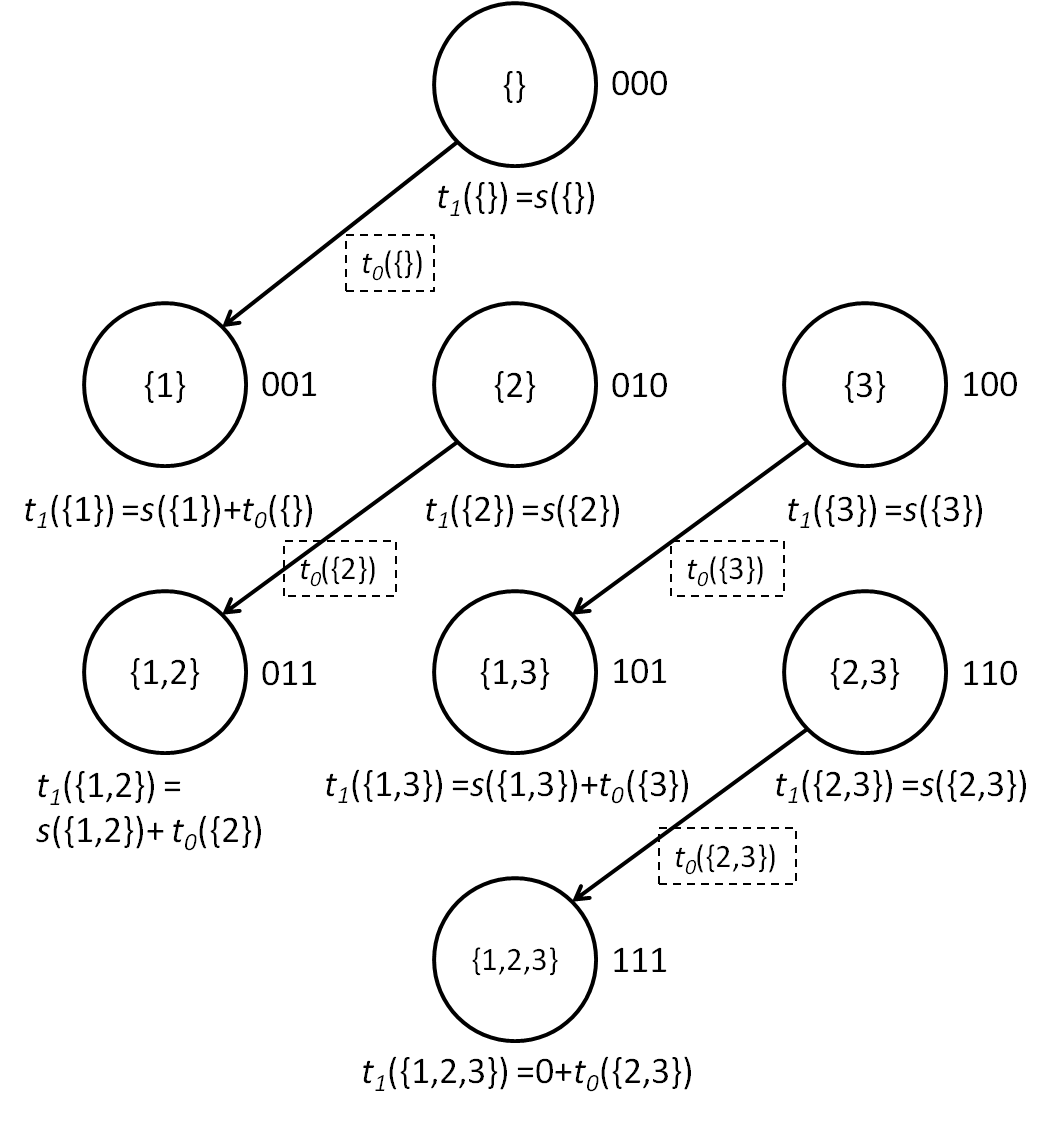}
		\caption{$j=1$}
	\end{subfigure}\\~\\
	\begin{subfigure}[b]{0.48\linewidth}
		\centering
		\includegraphics[width=0.95\linewidth]{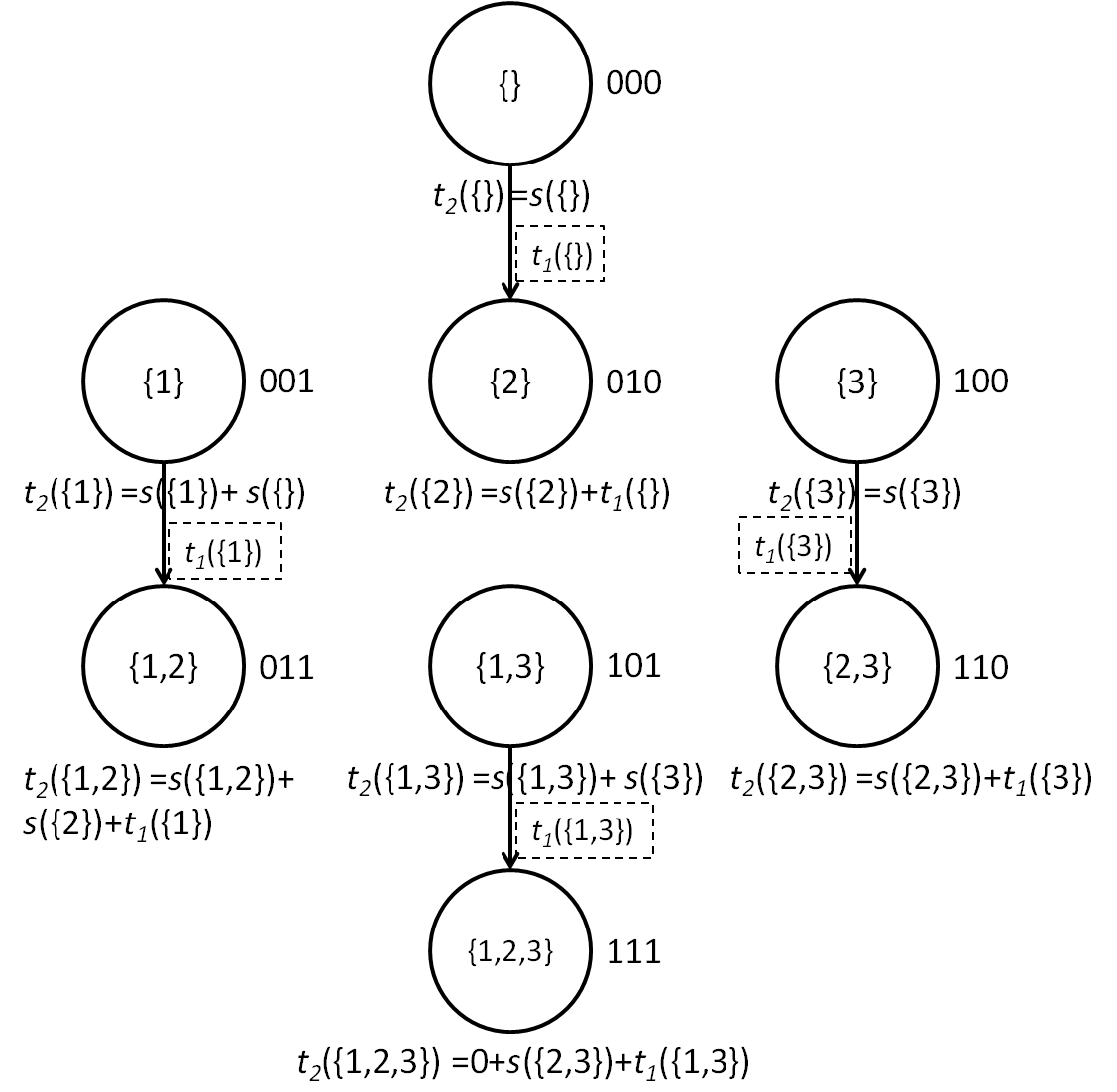}
		\caption{$j=2$}
	\end{subfigure}
	\begin{subfigure}[b]{0.48\linewidth}
		\centering
		\includegraphics[width=0.95\linewidth]{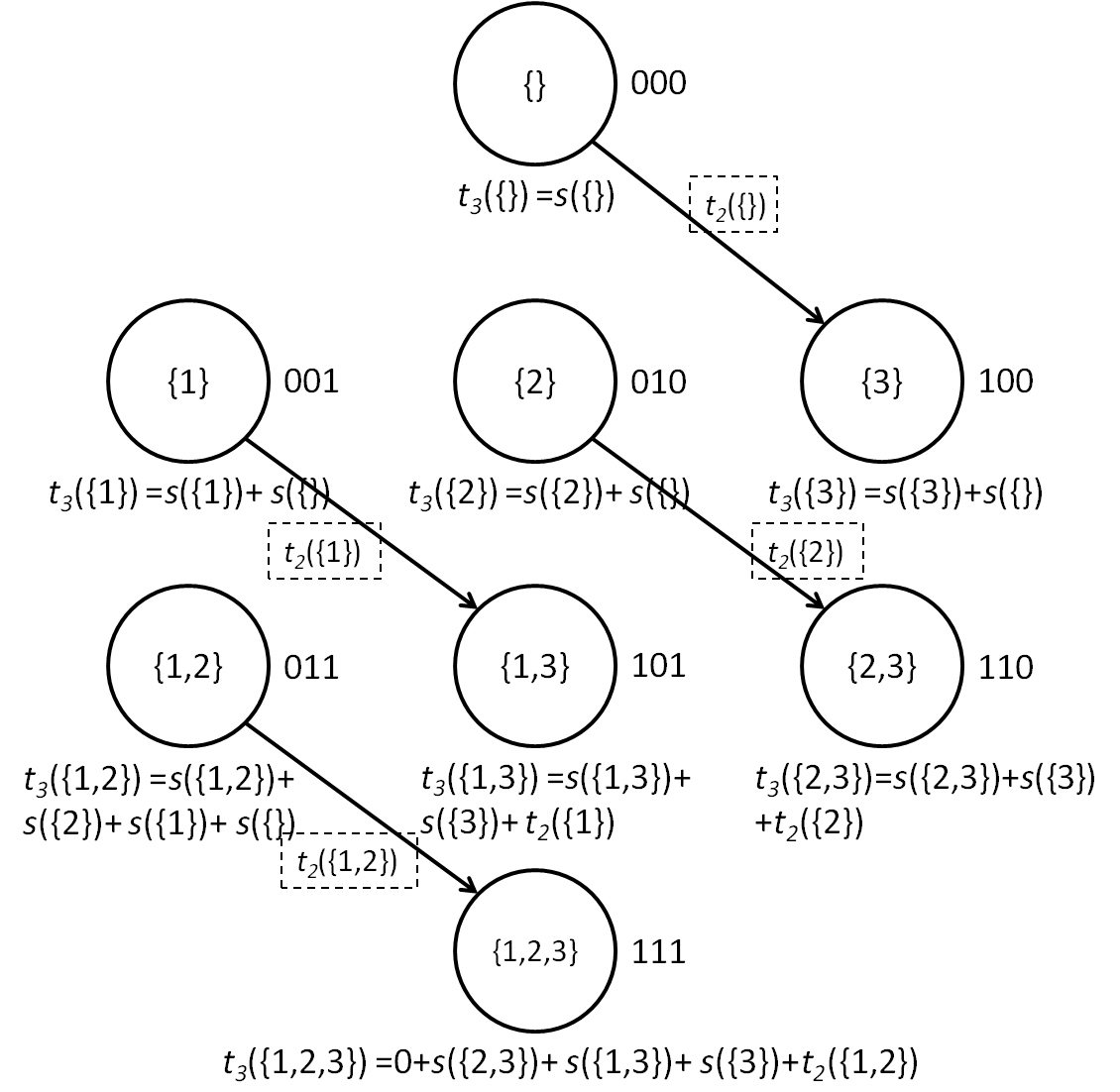}
		\caption{$j=3$}
	\end{subfigure}	
	\caption{An illustrative example of parallel truncated upward zeta transform on $n$-$D$ hypercube. In this case, $n=3$, $d=2$. The algorithm takes four iterations. 
		Iterations 0, 1, 2 and 3 are show in (a), (b), (c) and (d), respectively. The functions in dashed boxes are the messages sending between the processors. 
		In each iteration, the computed $t_j(S)$ is shown under each processor.}
	\label{fig:nd-upzeta} 
\end{figure}

\begin{figure}
	\centering
	\begin{subfigure}[b]{0.48\linewidth}
		\centering
		\includegraphics[width=0.9\linewidth]{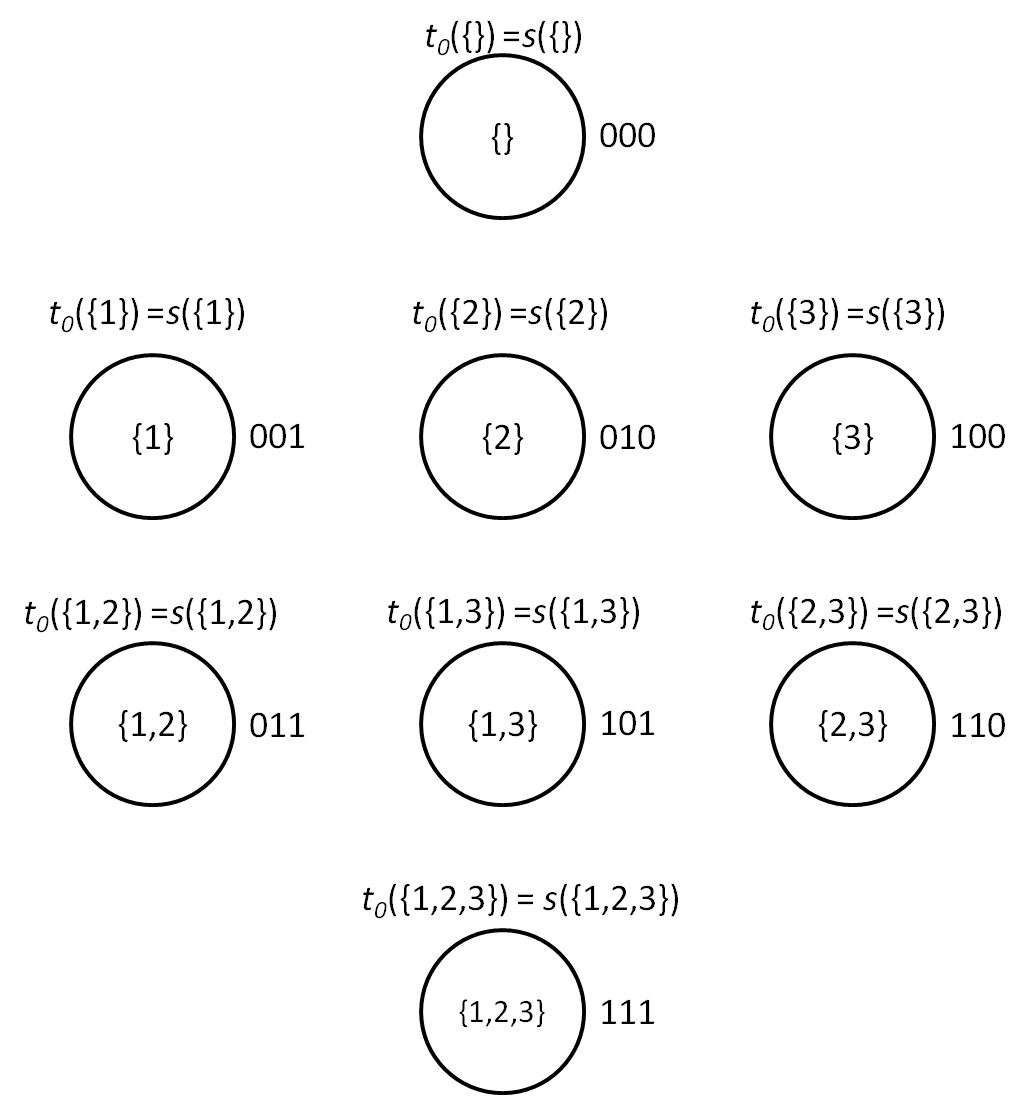}
		\caption{$j=0$}
	\end{subfigure}
	\begin{subfigure}[b]{0.48\linewidth}
		\centering
		\includegraphics[width=0.95\linewidth]{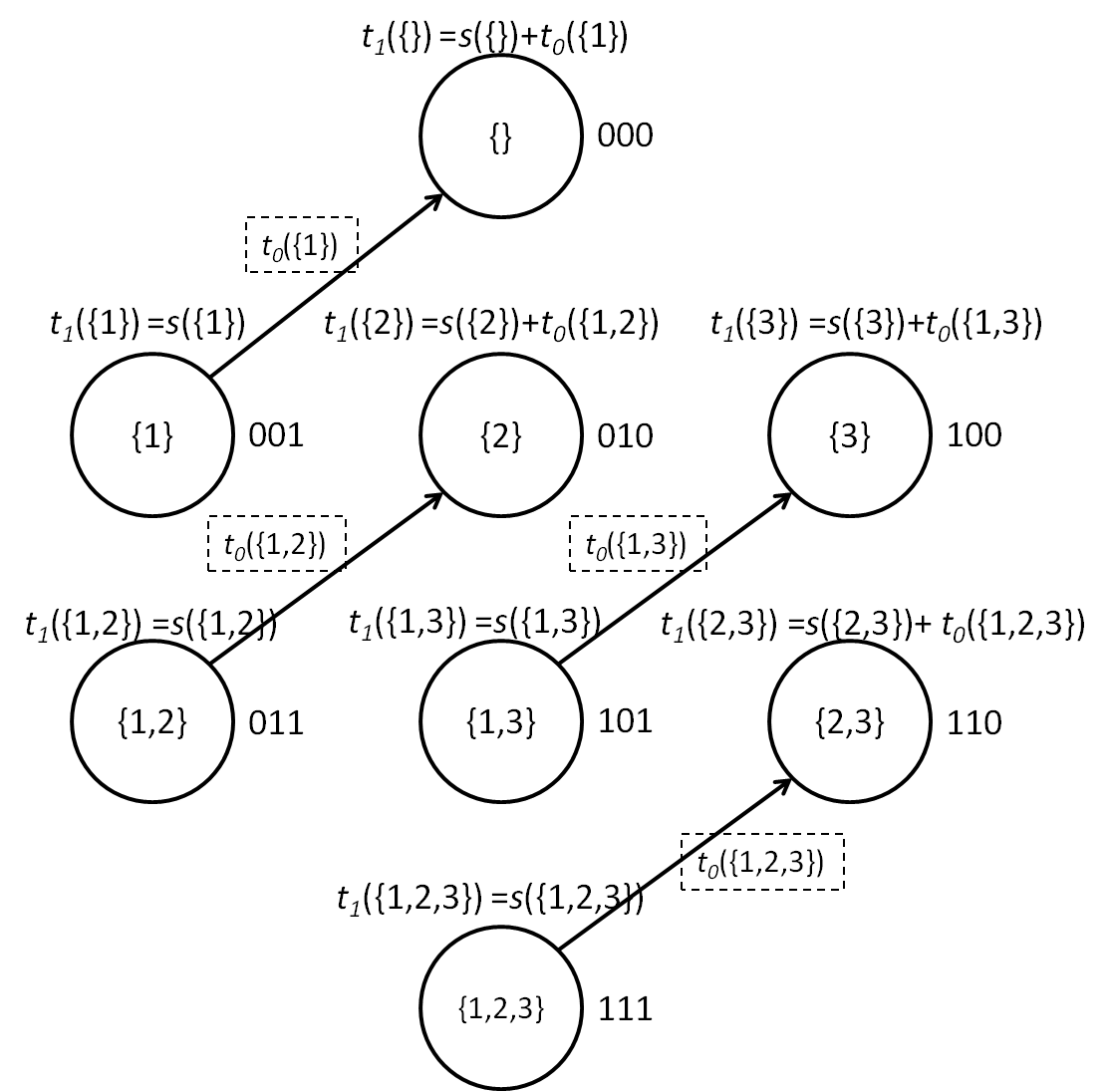}
		\caption{$j=1$}
	\end{subfigure}\\~\\
	\begin{subfigure}[b]{0.48\linewidth}
		\centering
		\includegraphics[width=0.95\linewidth]{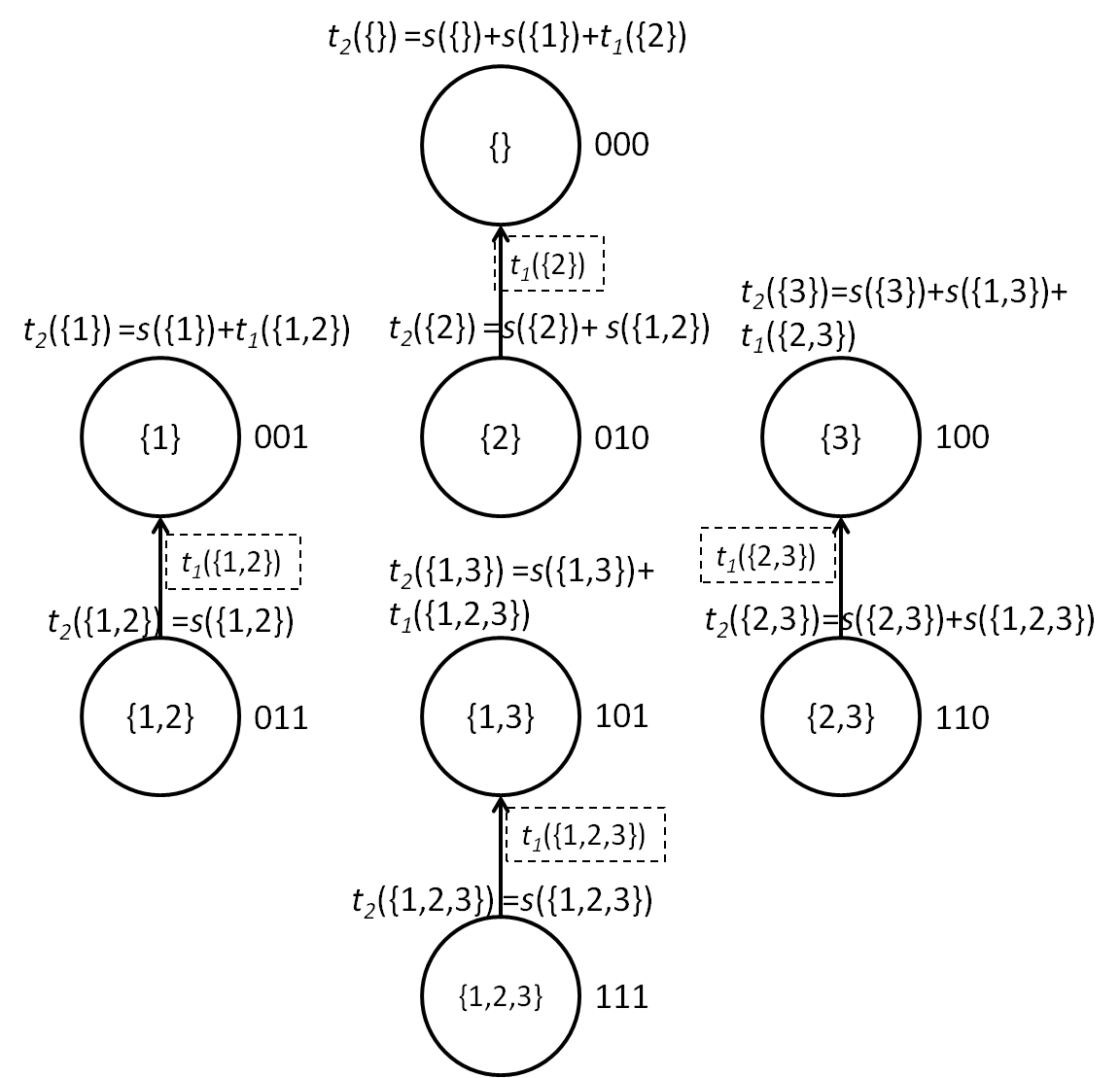}
		\caption{$j=2$}
	\end{subfigure}
	\begin{subfigure}[b]{0.48\linewidth}
		\centering
		\includegraphics[width=0.95\linewidth]{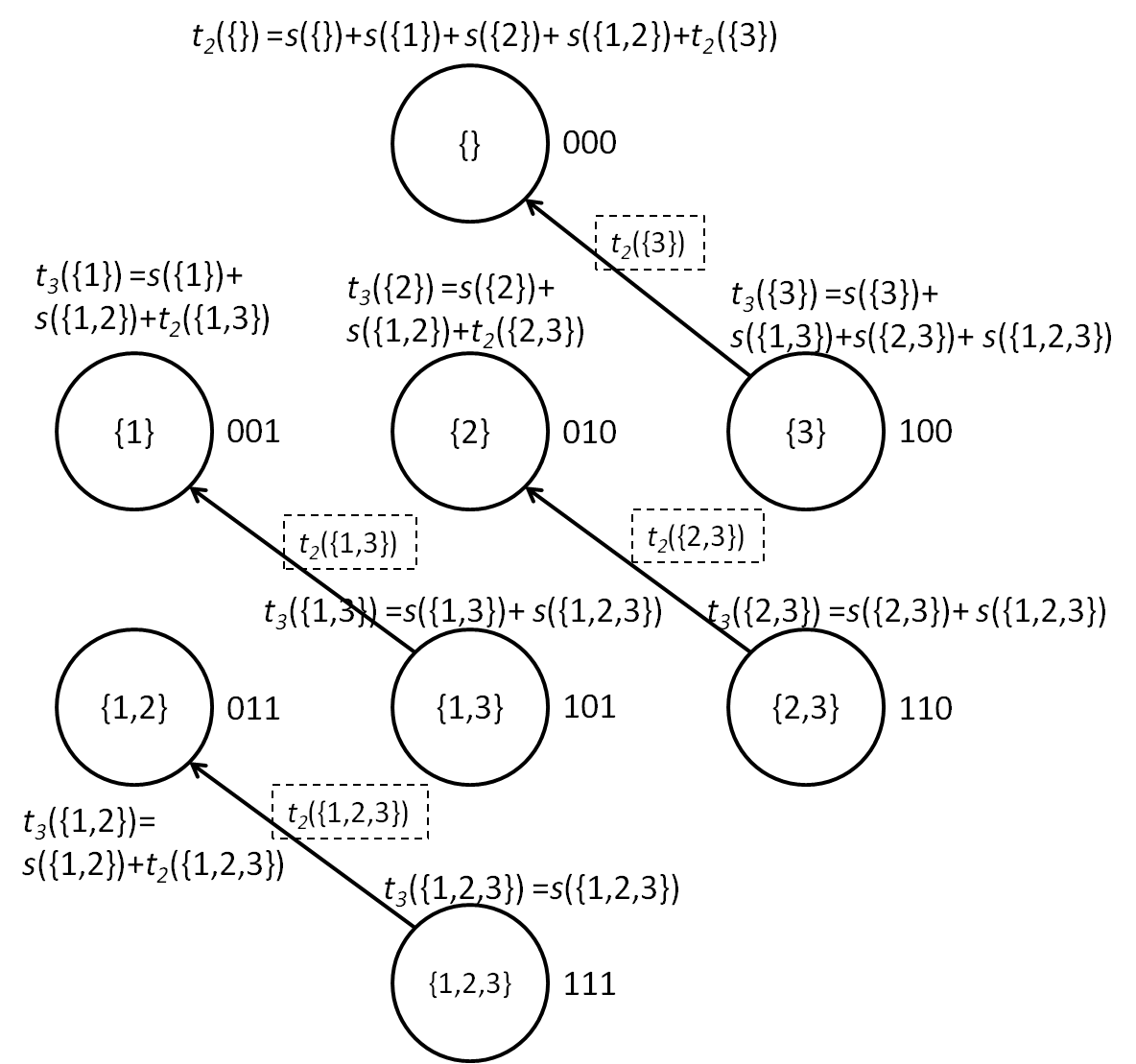}
		\caption{$j=3$}
	\end{subfigure}	
	\caption{An illustrative example of parallel truncated downward zeta transform on $n$-$D$ hypercube. In this case, $n=3$, $d=2$. The algorithm takes four iterations.
		 Iterations 0, 1, 2 and 3 are show in (a), (b), (c) and (d), respectively. The functions in dashed boxes are the messages sending between the processors. 
		In each iteration, the computed $t_j(S)$ is shown above each processor.}
	\label{fig:nd-downzeta} 
\end{figure}

By our definition in Section~3.1.1, a subset $S$ is encoded by an $n$-bit string $\omega$, where $\omega[i]=1$ if variable $i\in S$ and $\omega[i]=0$ otherwise. 
In an $n$-$D$ hypercube, $\omega$ is also used to denote the \emph{id} of a processor. 
We can take this natural mapping so that each processor $\omega$ is responsible for the corresponding subset $S$.
This forms the basic idea of the two parallel algorithms.

Algorithm~\ref{algorithm2} runs for $n+1$ iterations. In each iteration, all $2^n$ processors operate on their $S\subseteq V$ concurrently (Lines 3 to 12).
In iteration $j$, before the computation starts, each processor $\omega$ with $\omega[j]=0$ sends its $t_{j-1}(S)$ to its neighbor $\omega'$ obtained by inverting its $\omega[j]$ to 1, 
i.e., $\omega'=\omega\oplus2^{j-1}$.\footnote{$\oplus$ stands for the bitwise exclusive or (XOR) between two binary strings. 
$2^{j-1}$ stands for the binary string of integer $2^{j-1}$.} 
The neighbor receiving this $t_{j-1}$ will perform the addition on line 10 in iteration $j$ if necessary.
Figure~\ref{fig:nd-upzeta} illustrates an example of Algorithm~\ref{algorithm2} solving a problem with $n=3$ and $d=2$.

In Algorithm~\ref{algorithm3}, the mapping of the computation of $S$ to $n$-$D$ hypercube is the same as in Algorithm~\ref{algorithm2}. 
In iteration $j$, after the computation starts, each processor $\omega$ with $\omega[j]=1$ sends its $t_{j-1}$ to its neighbor $\omega'$ obtained by inverting its $\omega[j]$ to 0, i.e., $\omega'=\omega\oplus2^{j-1}$. The neighbor receiving this $t_{j-1}$ will perform the addition on line 7 in iteration $j$ if necessary.
Figure~\ref{fig:nd-downzeta} illustrates an example of Algorithm~\ref{algorithm3} solving a problem with $n=3$ and $d=2$.

In each iteration, all $S\subseteq V$ are computed concurrently on a $n$-$D$ hypercube, the computation times for both parallel algorithms being $O(n)$. 
Further, in both algorithms, the communications happen only between neighboring processors (two binary strings $\omega,\omega'$ differ in only one bit).
Thus, the two algorithms are communication-efficient.

We can use Algorithm~\ref{algorithm2} to compute $A_i(S)$ for a given $i\in V$ and all $S\subseteq V-\{i\}$ 
by setting $s(\cdot)=B_i(\cdot)$ and then computing $A_i(S)=q_i(S)\cdot t(S)$.\footnote{$A_i(S)$ are defined for all $S\subseteq V-\{i\}$, instead of  $S\subseteq V$. 
However, the algorithm can still be deployed by setting $t(S)=0$ for all $S\subseteq V$ s.t. $i\in S$.}
Note that $B_i(S)$ for any $S\subseteq V-\{i\}$ and $|S|\le d$ has also been computed on processor corresponding to $S$ since $s(S)=B_i(S)$.
To compute $A_i(S)$ for all $i\in V$, we run Algorithm~\ref{algorithm2} $n$ times with each time switching to the corresponding $q_i$ and $B_i$ functions.
Thus, $A_i$ for all $i\in V$ can be computed in $|V|\cdot O(n)=O(n^2)$ time. Each processor $\omega$ computes and keeps the corresponding $A_i(S)$ for all $i\in V$, which is the assumption we made in Section 3.1.1. Thus, the mapping adopted by the two algorithms is well suited for our algorithm as it avoids a large number of messages being passed when the computation transits to the next step.

\begin{figure}
	\centering
	\includegraphics[width=0.4\textwidth]{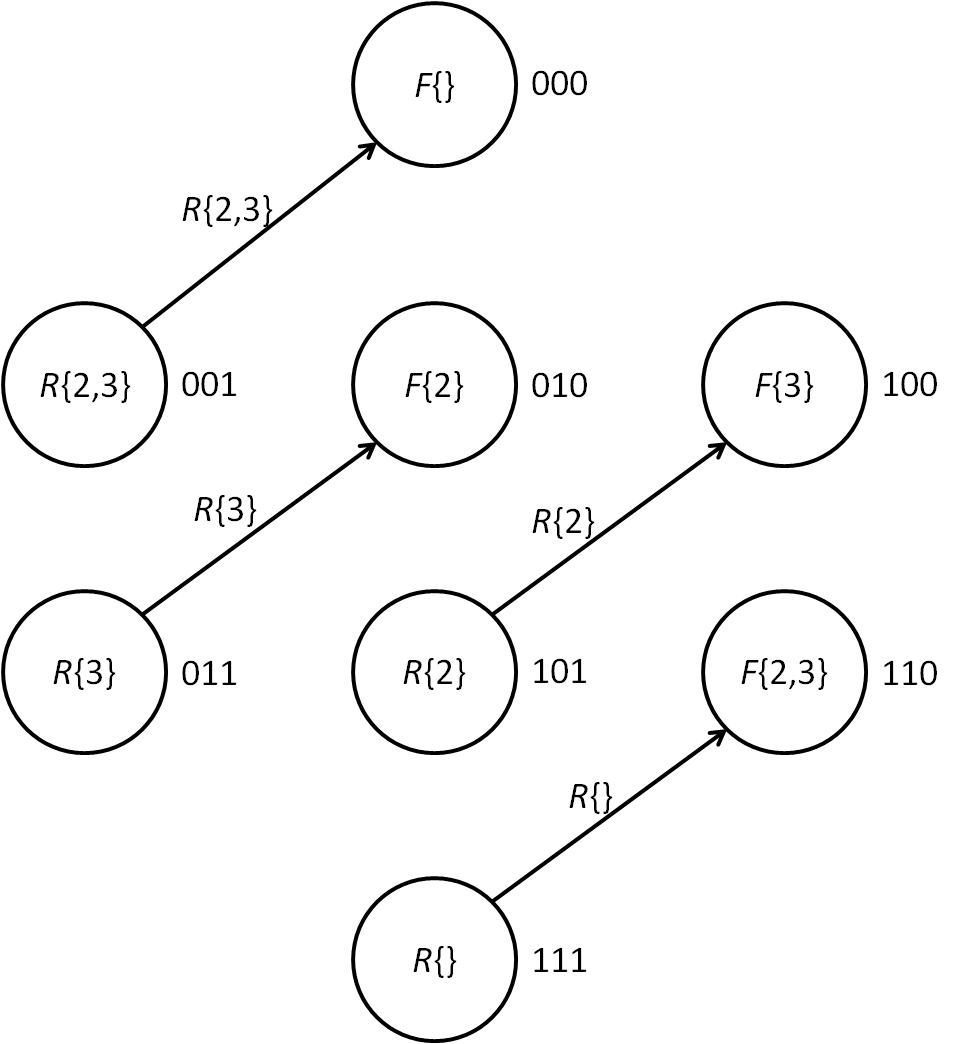}
	\caption{\small Retrieve $R$ for computing $\Gamma_v$. The example shows the case in which $\Gamma_v$ for $v=1$ is computed.}
	\label{fig3}
	\vspace{-1em}
\end{figure}

We will use Algorithm~\ref{algorithm3} to compute $\Gamma_v$ for a given $v\in V$. 
However, before applying the algorithm, we shall first compute $q_v(S)F(S)R(V-\{v\}-S)$ on the processor corresponding to $S$ (See Eq.~(12)). 
However, $F(S)$ and $R(V-\{v\}-S)$ are not on the same processor at the time when we have computed functions $F$ and $R$. Fortunately, they are on the processors who are neighbors in the hypercube. 
Thus, the processor $\omega$ who has $F(S)$ shall retrieve $R(V-\{v\}-S)$ from its neighbor $\omega'$ obtained by inverting its $\omega[v]$ to 1, i.e., $\omega'=\omega\oplus2^{v-1}$ (see example in Figure~\ref{fig3}), and compute $q_v(S)F(S)R(V-\{v\}-S)$ before Algorithm~\ref{algorithm3} is run. 
Then with Algorithm~\ref{algorithm3}, computing $\Gamma_v$ for any fixed $v\in V$ takes $O(n)$. 
The time for computing $\Gamma_v(G_v)$ for all $v\in V$ and $G_v\subseteq V-\{v\}$ and $|G_v|\le d$ is therefore $|V|\cdot O(n)=O(n^2)$.
 
\subsubsection{Computing $P(u\rightarrow v|D)$}
With $B_v(G_v)$ and $\Gamma_v(G_v)$ computed, we can compute $P(u\rightarrow v,D)$ using Eq. (11). 
Noting that $B_v(G_v)$ and $\Gamma_v(G_v)$ for any $G_v\subseteq V-\{v\}$ are on the same processor, each processor first computes $B_v(G_v)\cdot\Gamma_v(G_v)$ locally, 
then a \texttt{MPI\_Reduce}, a collective function in MPI library is executed on the hypercube to compute the sum of $B_v(G_v)\cdot\Gamma_v(G_v)$ from all processors. 
$P(u\rightarrow v|D)$ is then obtained by evaluating $P(u\rightarrow v, D)/F(V)$ at the processor with the highest rank, i.e., all bits in its \emph{id} are 1's. 
A \texttt{MPI\_Reduce} operation on a $n$-$D$ hypercube requires $O((\tau+\mu m)n)$ time, 
where $\tau$, $\mu$, $m$ are constants, specifying the latency, bandwidth of the communication network, and the message size.
Thus, computing $P(u\rightarrow v|D)$ for all $u,v\in V, u\neq v$ takes $O((\tau+\mu m)n^3)$ time. 

Adding up the time for each step, the time for evaluating all $n(n-1)$ edges is $O(n^3)$. As the sequential run-time is $O(2n(d+1)2^n)$, the parallel efficiency is $\Theta(2(d+1)/n^2)$.

\subsection{$k$-$D$ Hypercube Algorithm}
In Section 3.1, we have described the development of our parallel algorithm on an $n$-$D$ hypercube. However, we usually expect the number of processors $p\ll 2^n$.
Let $p=2^k$ be the number of processors, where $k<n$. We assume that the processors can communicate as in a $k$-$D$ hypercube. 
The strategy is to decompose the $n$-$D$ lattice
into $2^{n-k}$ $k$-$D$ lattices and map each $k$-$D$ lattice to the $p=2^k$ processors ($k$-$D$ hypercube). 

Following our previous definition, we use the binary string $\omega$ to denote the corresponding hypercube node $S$.
We number the positions of a binary string using $1,...,n$ (from right-most bit to left-most bit), and use $\omega[i,j]$ to denote the substring of $\omega$ between and including positions $i$ and $j$.
We partition the $n$-$D$ lattice into $2^{n-k}$ $k$-$D$ lattices based on the left $n-k$ bits of node \emph{id}'s. For a lattice node $\omega$, $\omega[k+1,n]$ specifies
the $k$-$D$ lattice it is part of and $\omega[1,k]$ specifies the \emph{id} of the processor it is assigned to. 
As an example, Figure~\ref{decompose} shows the decomposition of an 3-$D$ lattice to two 2-$D$ lattices and the mapping to an 2-$D$ hypercube computing cluster.
In this case, subsets \texttt{$\{\}$} and \texttt{$\{3\}$} are assigned to processor \texttt{00}, \texttt{$\{1\}$} and \texttt{$\{1,3\}$} are assigned to processor \texttt{01}, so on and so forth.
Thus, each processor in a $k$-$D$ hypercube is responsible for computing relevant scores for $2^{n-k}$ $S$ subsets. This forms the basic idea of our $k$-$D$ hypercube algorithm.

\begin{figure}[h!]
\centering
\includegraphics[width=1.0\textwidth]{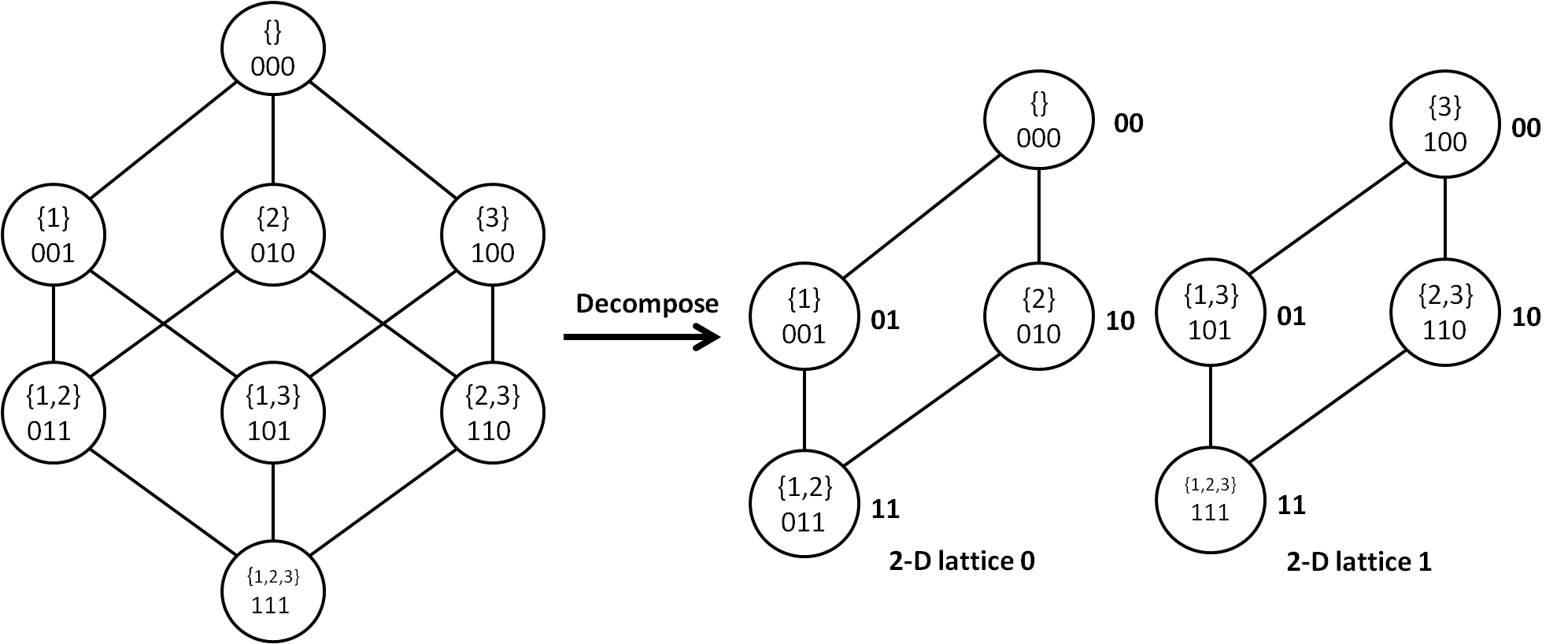}
\caption{\small Decompose a 3-$D$ lattice into two 2-$D$ lattices which are then mapped to an 2-$D$ hypercube. 
The 3-bit binary string inside the node represents the binary code of the corresponding subset $S$. The 2-bit binary string beside each node denotes the \emph{id} of the processor in the 2-$D$ hypercube.}
\label{decompose}
\end{figure}

In the following, we first develop $k$-$D$ hypercube algorithms for the two \emph{zeta} transform variants. 
We then present the $k$-$D$ hypercube algorithms for computing $F$ and $R$ functions.
Finally we introduce the overall $k$-$D$ hypercube algorithm for computing the edge posteriors.

\subsubsection{Parallel Fast Zeta Transforms on $k$-$D$ hypercube}

In order to compute $A$ and $\Gamma$ functions on a $k$-$D$ hypercube, we generalize Algorithms \ref{algorithm2} and \ref{algorithm3}.
We number the processors in the $k$-$D$ hypercube computer cluster with a $k$-bit binary string $r$ such that two adjacent processors $r$, $r'$ differ in one bit. 
The basic idea is, instead of computing the transform for only one subset $S$,
each processor $r$ is responsible for computing $2^{n-k}$ subsets $S$ such that $r=\omega[1,k]$.
We present the generalized algorithms for the two transforms in Algorithm~\ref{algorithm4} and Algorithm~\ref{algorithm5}, respectively.

Figure~\ref{fig:kd-upzeta} shows a running example of Algorithm~\ref{algorithm4} with $n=3$, $d=2$ and $k=2$.
In this case, we have 8 subsets and each processor is computing two subsets. 
Another notable difference from the example in Figure~\ref{fig:nd-upzeta} is that when $j>k$, $t_{j-1}(S-\{j\})$'s are available locally 
thus no message passing between processor is needed. 
Similarly, Figure~\ref{fig:kd-downzeta} shows a running example of Algorithm~\ref{algorithm5} with $n=3$, $d=2$ and $k=2$. 

We now present two theorems that respectively characterize the run-time complexities of the two algorithms.
\begin{theorem}
	\label{kdupzetatheory}
	Algorithm~\ref{algorithm4} computes the truncated upward zeta transform in time $O((d+1)\cdot 2^{n-k}+k(n-k)^d)$ on $k$-$D$ hypercube.
\end{theorem}

\begin{proof}
	As it is specified, each processor $r$ computes subsets $S$ s.t. $r=\omega[1,k]$.
	Algorithm~\ref{algorithm4} runs for $n$ iterations. For the iterations $j=n-d, ..., n$, all $S\subseteq V$ satisfy the condition on line 3, thus each processor
	performs the computation on lines 3--14 for the corresponding $2^{n-k}$ subsets on it. The total computing time for these iterations is $O((d+1)2^{n-k})=O(d2^{n-k})$.
	
	For iterations $j=1,...,n-d-1$, the processor $r$ s.t. $r[i]=0$ for all $i=1,..,k$ has the largest number of subset $S$ that satisfy the condition on line 3, thus computes lines 4--13 the most frequently among all the processors. The computation time of the algorithm for these iterations is up-bounded by its computing time. Thus, for iterations $j=1,...,n-d-1$, we only need to characterize this processor's computing time, which is proportional to
	\begin{flalign}
		\begin{split}
			&\sum\limits_{j=1}^k\sum\limits_{r=0}^d{n-k\choose r}+\sum\limits_{j=k+1}^{n-d-1}2^{j-k}\sum\limits_{r=0}^d{n-j\choose r}
			= k\sum\limits_{r=0}^d{n-k\choose r} + 2^{-k}\sum\limits_{j=k+1}^{n-d-1}2^j\sum\limits_{r=0}^d{n-j\choose r}\\
			&\leq k\sum\limits_{r=0}^d{n-k\choose r} + 2^{-k}\sum\limits_{j=k+1}^{n-d-1}2^j(n-j)^d
			\leq k\sum\limits_{r=0}^d{n-k\choose r} + 2^{-k}\sum\limits_{j=1}^{n-d-1}2^j(n-j)^d\\
			&\leq k\sum\limits_{r=0}^d{n-k\choose r} + 2^{-k}2^n\sum\limits_{j=0}^{\infty}(1/2)^jj^d
		\end{split}
	\end{flalign} 
	The first term $k\sum\limits_{r=0}^d{n-k\choose r} =O(k(n-k)^d)$. The second term 
	$2^{-k}2^n\sum\limits_{j=0}^{\infty}(1/2)^jj^d=O(2^{n-k})$ as the infinite sum converges to a finite limit for a fixed $d$. 
	Thus, the time combined for all iteration is $O(k(n-k)^d)+O(2^{n-k})+O(d2^{n-k})=O((d+1)\cdot 2^{n-k}+k(n-k)^d)$.
\end{proof}

\begin{algorithm}\small
	\caption{Parallel Truncated Upward Zeta Transform on $k$-$D$ hypercube}
	\label{algorithm4}
	\begin{algorithmic}[1]
		\Assumption $1\le k\le n$, each subset $S\subseteq V$ is encoded by an $n$-bit binary string $\omega$, where $\omega[i]=1$ if variable $i\in S$ and $\omega[i]=0$ otherwise. Each processor in the $k$-$D$ hypercube is encoded by an $k$-bit binary string $r$. Subset $S$ is computed on processor $r=\omega[1,k]$.
		\State On each processor $r$, $t_0(S)\leftarrow s(S)$ for all $|S|\leq d$ s.t. $r=\omega[1,k]$ and $t_0(S)\leftarrow 0$ otherwise.
		\For{$j\leftarrow 1$ to $n$}
		\For{each $S\subseteq V$ with $|S\cap\{j+1,...,n\}|\leq d$ on each processor $r$}
		\State $t_j(S)\leftarrow 0$
		\If{$|S\cap \{j,...,n\}|\leq d$}
		\State $t_j(S)\leftarrow t_{j-1}(S)$
		\EndIf
		\If{$j\in S$}
		\If {$j\le k$} 
		\State Retrieve $t_{j-1}(S-\{j\})$ from processor $r'=r\oplus2^{j-1}$.
		\EndIf 
		\State $t_j(S)\leftarrow t_j(S)+t_{j-1}(S-\{j\})$
		\EndIf
		\EndFor
		\EndFor
		\State \textbf{return} $t_n(S)$
	\end{algorithmic}
\end{algorithm}

\begin{figure}
	\centering
	\begin{subfigure}[b]{0.48\linewidth}
		\centering
		\includegraphics[width=0.8\linewidth]{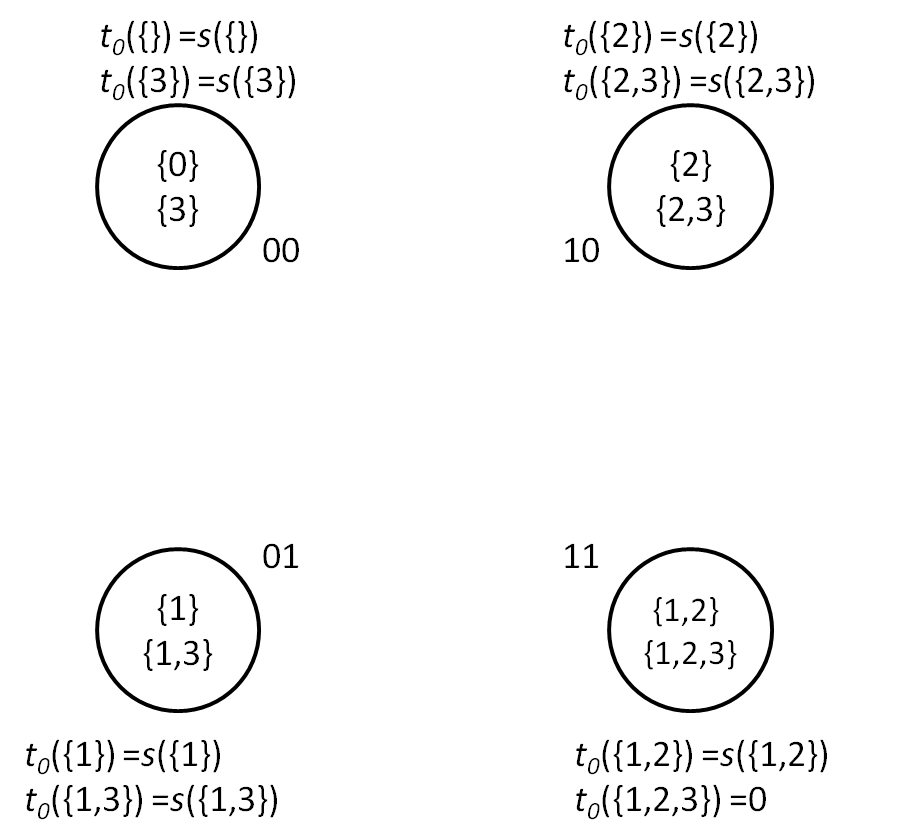}
		\caption{$j=0$}
	\end{subfigure}
	\begin{subfigure}[b]{0.48\linewidth}
		\centering
		\includegraphics[width=0.8\linewidth]{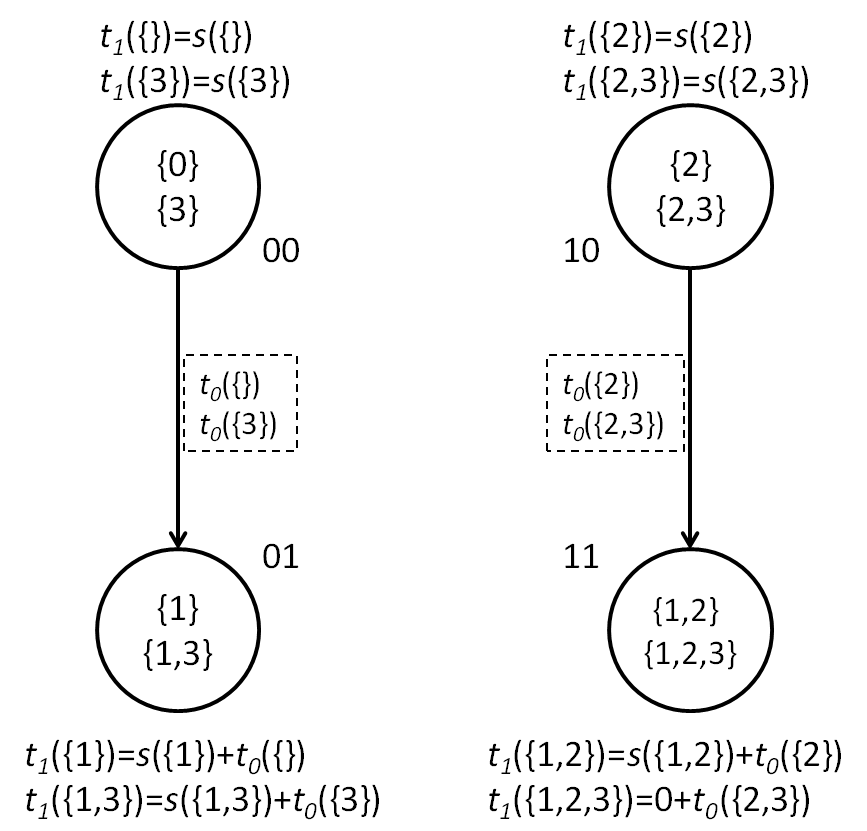}
		\caption{$j=1$}
	\end{subfigure}\\~\\
	\begin{subfigure}[b]{0.48\linewidth}
		\centering
		\includegraphics[width=0.85\linewidth]{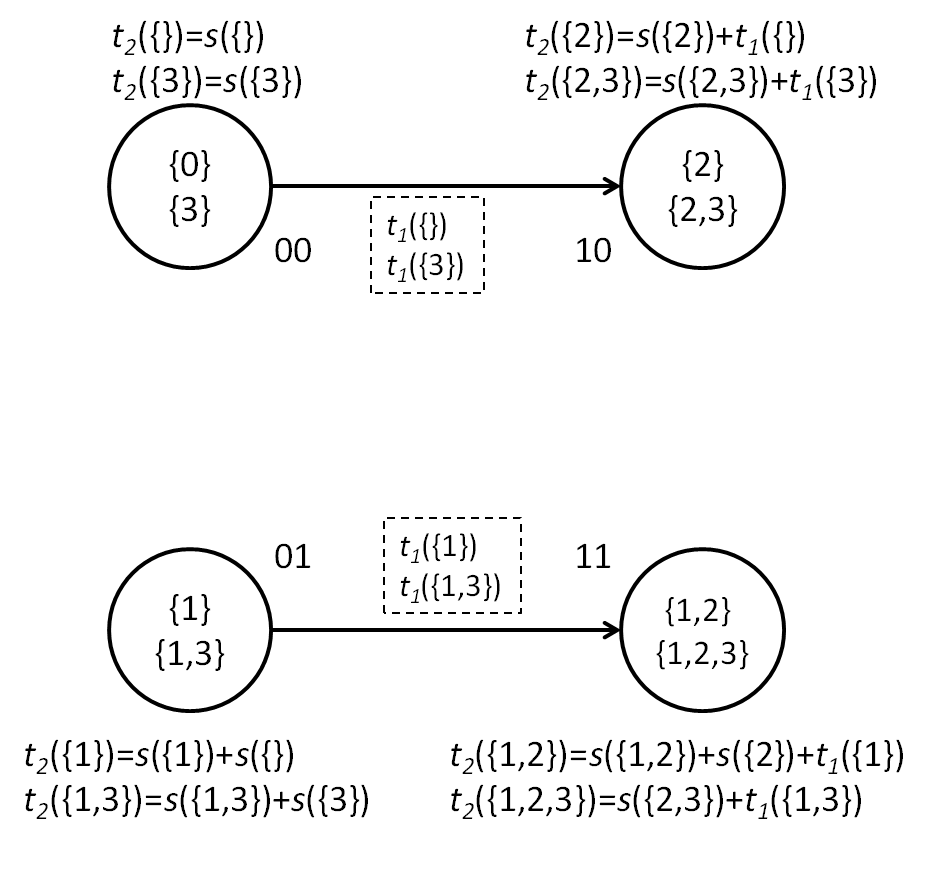}
		\caption{$j=2$}
	\end{subfigure}
	\begin{subfigure}[b]{0.48\linewidth}
		\centering
		\includegraphics[width=0.9\linewidth]{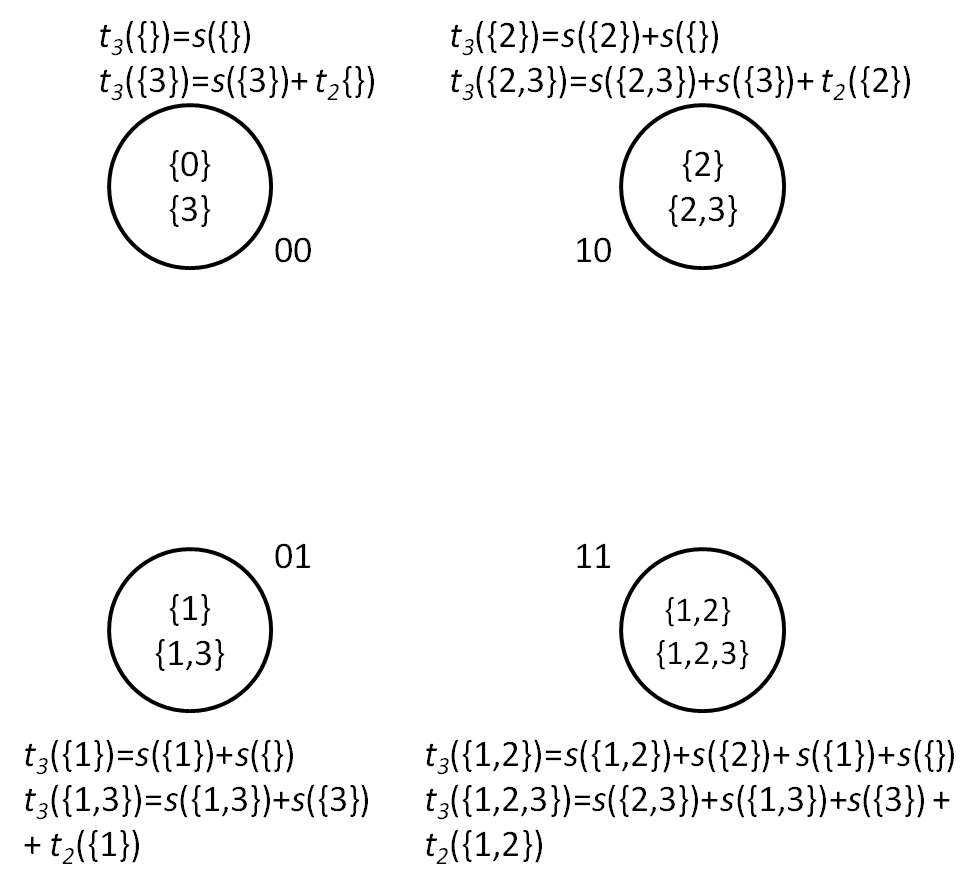}
		\caption{$j=3$}
	\end{subfigure}	
	\caption{\small An illustrative example of parallel truncated upward zeta transform on $k$-$D$ hypercube. In this case, $n=3$, $d=2$, $k=2$. The algorithm takes four iterations. 
		Iterations 0, 1, 2 and 3 are show in (a), (b), (c) and (d), respectively. The functions in dashed boxes are the messages sending between the processors. 
		In each iteration, the computed $t_j(S)$'s is shown under each processor. In (d), $j=3>k=2$, $t_{j-1}(S-\{j\})$'s are available locally thus no message passing is needed.}
	\label{fig:kd-upzeta} 
\end{figure}

\clearpage

\begin{algorithm}\small
	\caption{Parallel Truncated Downward Zeta Transform on $k$-$D$ hypercube}
	\label{algorithm5}
	\begin{algorithmic}[1]
		\Assumption $1\le k\le n$, each subset $S\subseteq V$ is encoded by an $n$-bit binary string $\omega$, where $\omega[i]=1$ if variable $i\in S$ and $\omega[i]=0$ otherwise. Each processor in the $k$-$D$ hypercube is encoded by an $k$-bit binary string $r$. Subset $S$ is computed on processor $r=\omega[1,k]$.
		\State On each processor, $t_0(S)\leftarrow s(S)$ for all $S$ s.t. $r=\omega[1,k]$.
		\For{$j\leftarrow 1$ to $n$}
		\For{each $S\subseteq V$ with $|S\cap\{1,...,j\}|\leq d$ on each processor $r$}
		\State $t_j(S)\leftarrow t_{j-1}(S)$
		\If{$j\notin S$}
		\If {$j\le k$} 
		\State Retrieve $t_{j-1}(S\cup\{j\})$ from processor $r'=r\oplus2^{j-1}$.
		\EndIf 
		\State $t_j(S)\leftarrow t_j(S)+t_{j-1}(S\cup\{j\})$
		\EndIf
		\EndFor
		\EndFor
		\State \textbf{return} $t_n(S)$
	\end{algorithmic}
	\vspace{-0.5em}
\end{algorithm}

\begin{theorem}
\label{kddownzetatheory}
Algorithm~\ref{algorithm5} computes the truncated downward zeta transform in time $O((4d+4)\cdot 2^{n-k})$ on $k$-$D$ hypercube.
\end{theorem}

\begin{proof}
Each processor $r$ computes subsets $S$ s.t. $r=\omega[1,k]$.
In Algorithm~\ref{algorithm5}, line 1 takes $O(2^{n-k})$ time.
Lines 2--12 runs for $n$ iterations. 
For the iterations $j=1, ..., d$, all $S\subseteq V$ satisfy the condition on line 3, thus each processor
performs the computation on line 4--10 for all $2^{n-k}$ subsets on it. Thus the total computation time for these iterations is $O(d2^{n-k})$.

For iterations $j=d+1,...,n$, the processor $r$ s.t. $r[i]=0$ for all $i=1,..,k$ enters the loop 3--11 more frequently than any other processor,
thus requires the most computation time. 
The running time of Algorithm~\ref{algorithm5} for these iterations is up-bounded by its running time, which is proportional to



\begin{flalign}
\vspace{-1em}
\begin{split}
&{\sum\limits_{j=d+1}^{k+d}2^{n-k}+\sum\limits_{j=k+d+1}^n2^{n-j}\sum\limits_{r=0}^d{j-k\choose r}=k2^{n-k}+2^{-k}\sum\limits_{i=d+1}^{n-k}2^{n-i}\sum\limits_{r=0}^d{i\choose r}}\\
&=k2^{n-k}+2^{-k}\sum\limits_{i=d+1}^{4d-1}2^{n-i}\sum\limits_{r=0}^d{i\choose r}+2^{-k}\sum\limits_{i=4d}^n2^{n-i}\sum\limits_{r=0}^d{i\choose r}-2^{-k}\sum\limits_{i=n-k+1}^{n}2^{n-i}\sum\limits_{r=0}^d{i\choose r}\\
&\leq k2^{n-k}+2^{-k}\sum\limits_{i=d+1}^{4d-1}2^n+2^{-k}\sum\limits_{i=4d}^{n-k}2^{n-i}\sum\limits_{r=0}^d{i\choose r}-2^{-k}\sum\limits_{i=n-k+1}^{n}2^{n-d}\sum\limits_{r=0}^d{d\choose r}\\
&= k2^{n-k}+(3d-1)2^{n-k}+2^{-k}\sum\limits_{i=4d}^{n-k}2^{n-i}\sum\limits_{r=0}^d{i\choose r}-2^{-k}\sum\limits_{i=n-k+1}^{n}2^{n-d}2^d\\
&=(k+3d-1)2^{n-k}+2^{-k}\sum\limits_{i=4d}^n2^{n-i}\sum\limits_{r=0}^d{i\choose r}-k2^{n-k}\\
&\leq (3d-1)2^{n-k}+5\cdot2^{n-k}
\end{split}
\vspace{-1em}
\end{flalign} 

The upper bound $2^{-k}\sum\limits_{i=4d}^n2^{n-i}\sum\limits_{r=0}^d{i\choose r}\leq 5\cdot2^{n-k}$ in last step is from \textbf{Corollary 3} in \citep{Koivisto06}. Thus, the run-time is $O((4d+4)\cdot 2^{n-k})$. 
\end{proof}

\begin{figure}[ht]
	\centering
	\begin{subfigure}[b]{0.48\linewidth}
		\centering
		\includegraphics[width=0.8\linewidth]{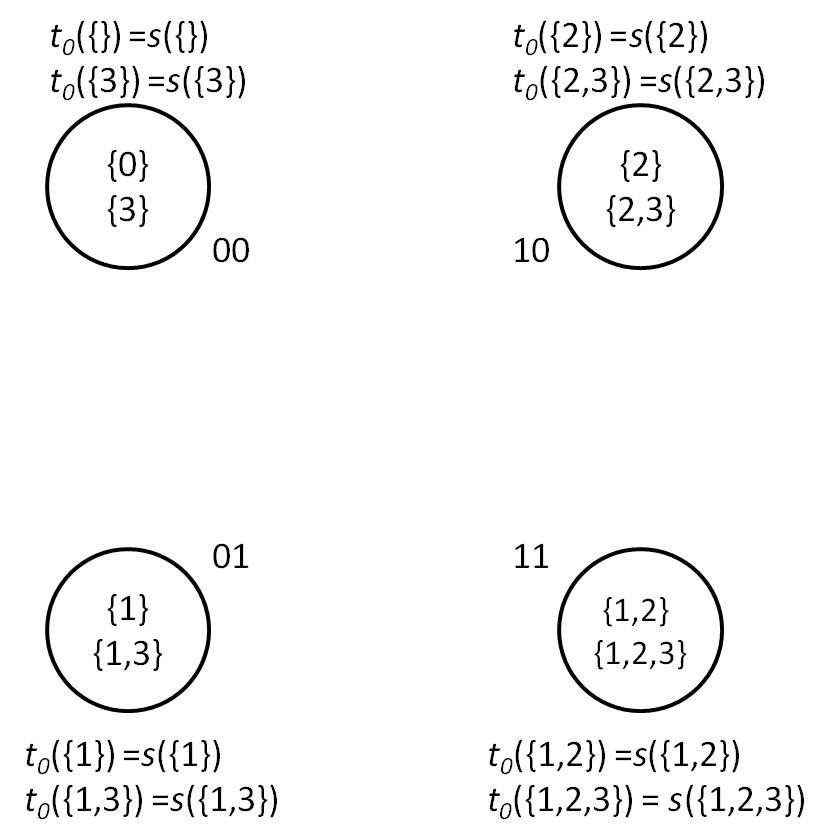}
		\caption{$j=0$}
	\end{subfigure}
	\begin{subfigure}[b]{0.48\linewidth}
		\centering
		\includegraphics[width=0.8\linewidth]{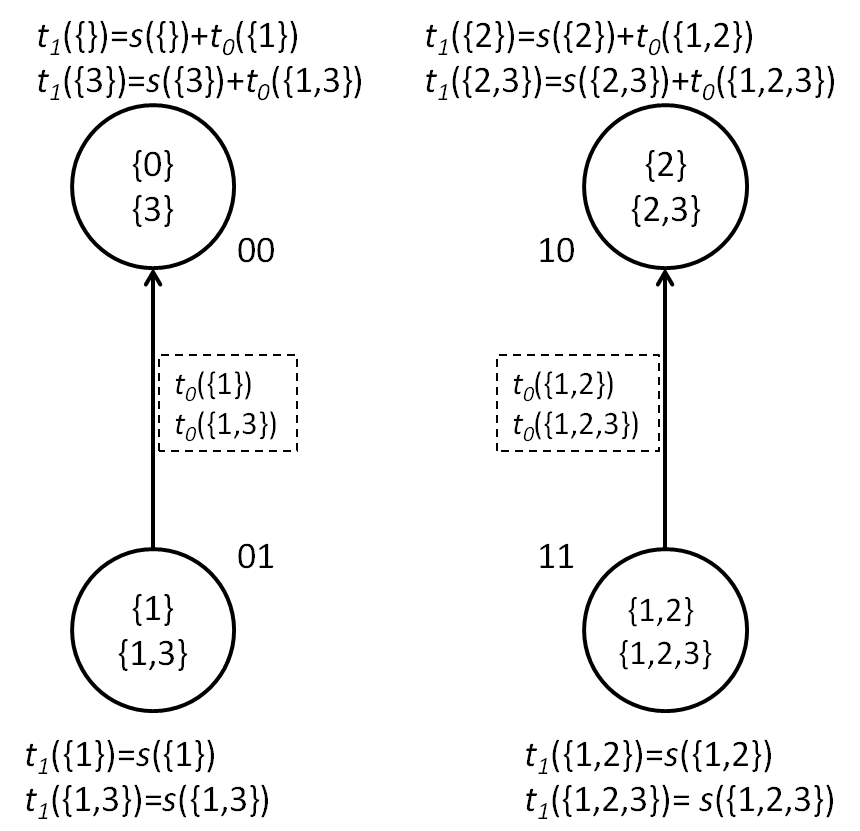}
		\caption{$j=1$}
	\end{subfigure}\\~\\
	\begin{subfigure}[b]{0.48\linewidth}
		\centering
		\includegraphics[width=0.9\linewidth]{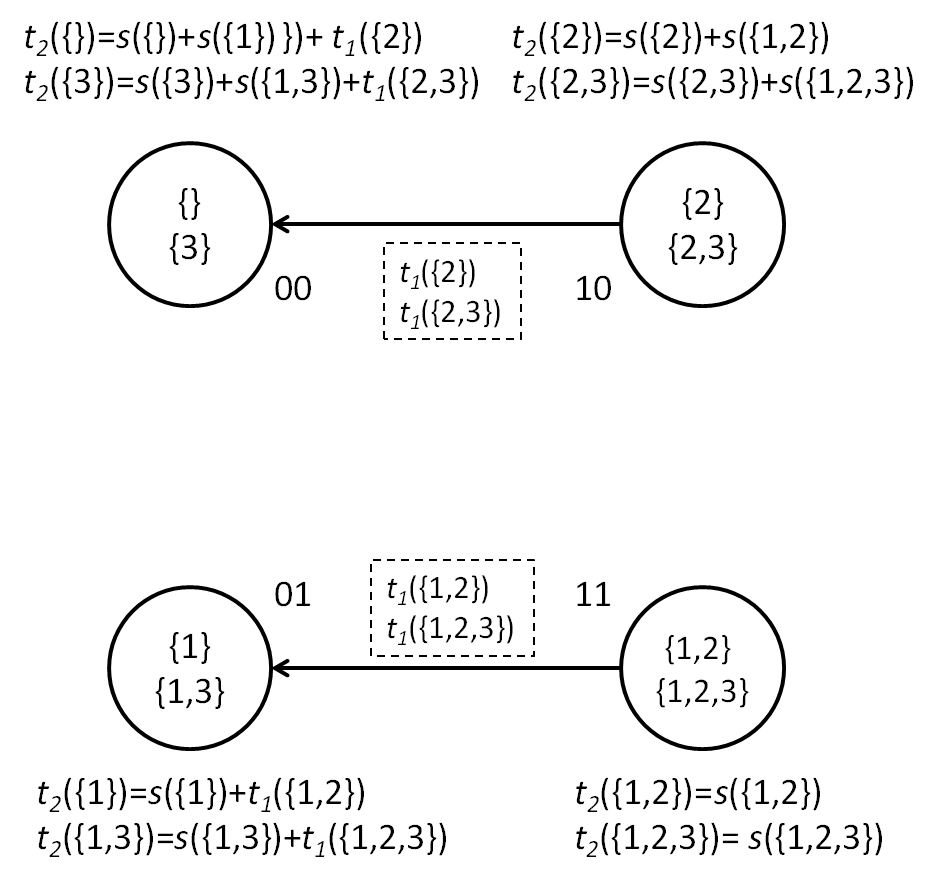}
		\caption{$j=2$}
	\end{subfigure}
	\begin{subfigure}[b]{0.48\linewidth}
		\centering
		\includegraphics[width=0.95\linewidth]{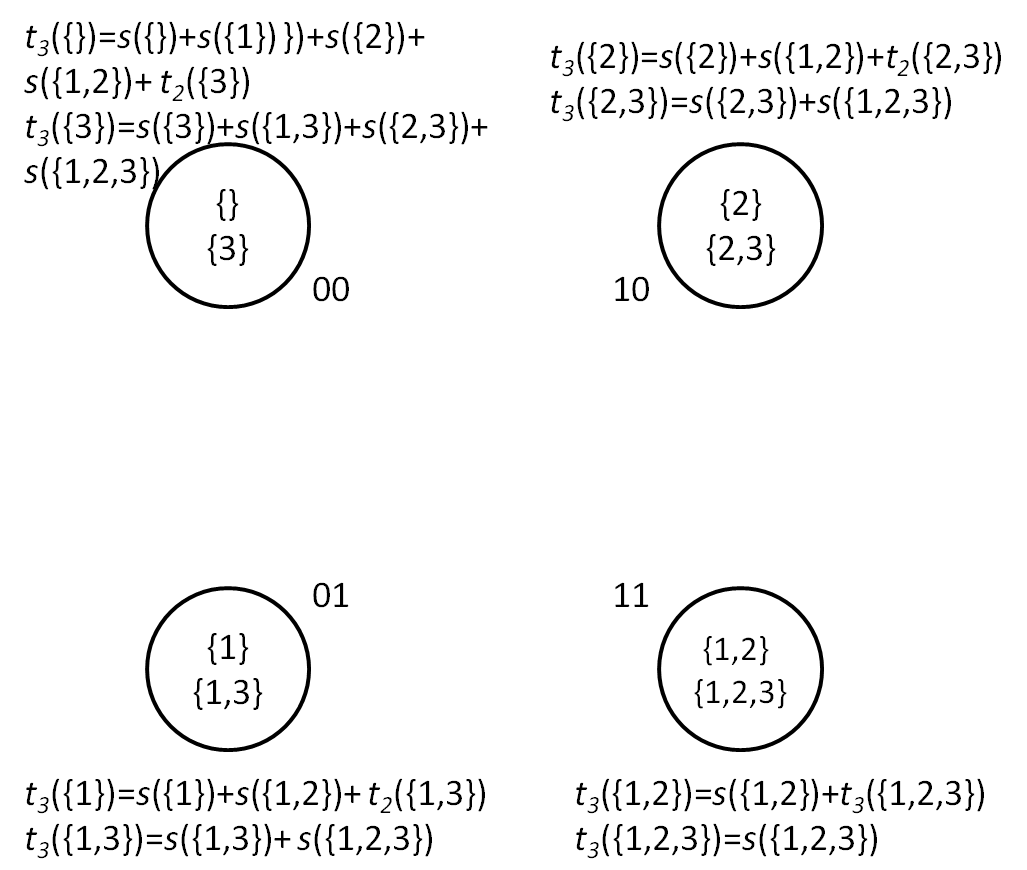}
		\caption{$j=3$}
	\end{subfigure}	
	\caption{\small An illustrative example of parallel truncated downward zeta transform on $k$-$D$ hypercube. In this case, $n=3$, $d=2$, $k=2$. The algorithm takes four iterations. 
		Iterations 0, 1, 2 and 3 are show in (a), (b), (c) and (d), respectively. The functions in dashed boxes are the messages sending between the processors. 
		In each iteration, the computed $t_j(S)$'s is shown under each processor. In (d), $j=3>k=2$, $t_{j-1}(S\cup\{j\})$'s are available locally thus no message passing is needed.}
	\label{fig:kd-downzeta} 
\end{figure}

\subsubsection{Computing $F(S)$ and $R(S)$ on $k$-$D$ Hypercube}

To compute function $F$, we partition the $n$-$D$ DP lattice into $2^{n-k}$ $k$-$D$ hypercubes based on the left $n-k$ bits of node \emph{id}'s. For a lattice node $\omega$, $\omega[k+1,n]$ specifies
the $k$-$D$ hypercube it is part of and $\omega[1,k]$ specifies the processor it is assigned to. Using the strategy proposed by \citep{nikolova2009parallel}, we pipeline the execution of the $k$-$D$ hypercubes to complete the parallel execution in $2^{n-k}+k$ time steps such that all processors are active
except for the first $k$ and last $k$ time steps during the buildup and finishing off of the pipeline. Specifically, let each $k$-$D$ hypercube denoted by an $(n-k)$ bit string,
which is the common prefix to the $2^k$ lattice/$k$-$D$ hypercube nodes that are part of this $k$-$D$ sub-hypercube. The $k$-$D$ hypercubes are processed in the increasing order
of the number of 1's in their bit string specifications, and in lexicographic order within the group of hypercubes with the same number of 1's. Formally, we have the following rule:
let $H_i$ and $H_j$ be two $k$-$D$ hypercubes and let $\omega_S$ and $\omega_T$ be the binary strings of two nodes $S$ and $T$ in the lattice that map to $H_i$ and $H_j$, respectively. Then, the computation of
$H_i$ is initiated before computation of $H_j$ if and only if:

\begin{enumerate}
	\item $\mu(\omega_S[k+1,n])<\mu(\omega_T[k+1,n])$, or
	\item $\mu(\omega_S[k+1,n])=\mu(\omega_T[k+1,n])$ and $\omega_S[k+1,n]$ is lexicographically smaller than $\omega_T[k+1,n]$.
\end{enumerate}

\begin{figure}
	\centering
	\begin{subfigure}[b]{0.49\linewidth}
		\centering
		\includegraphics[width=1.0\linewidth]{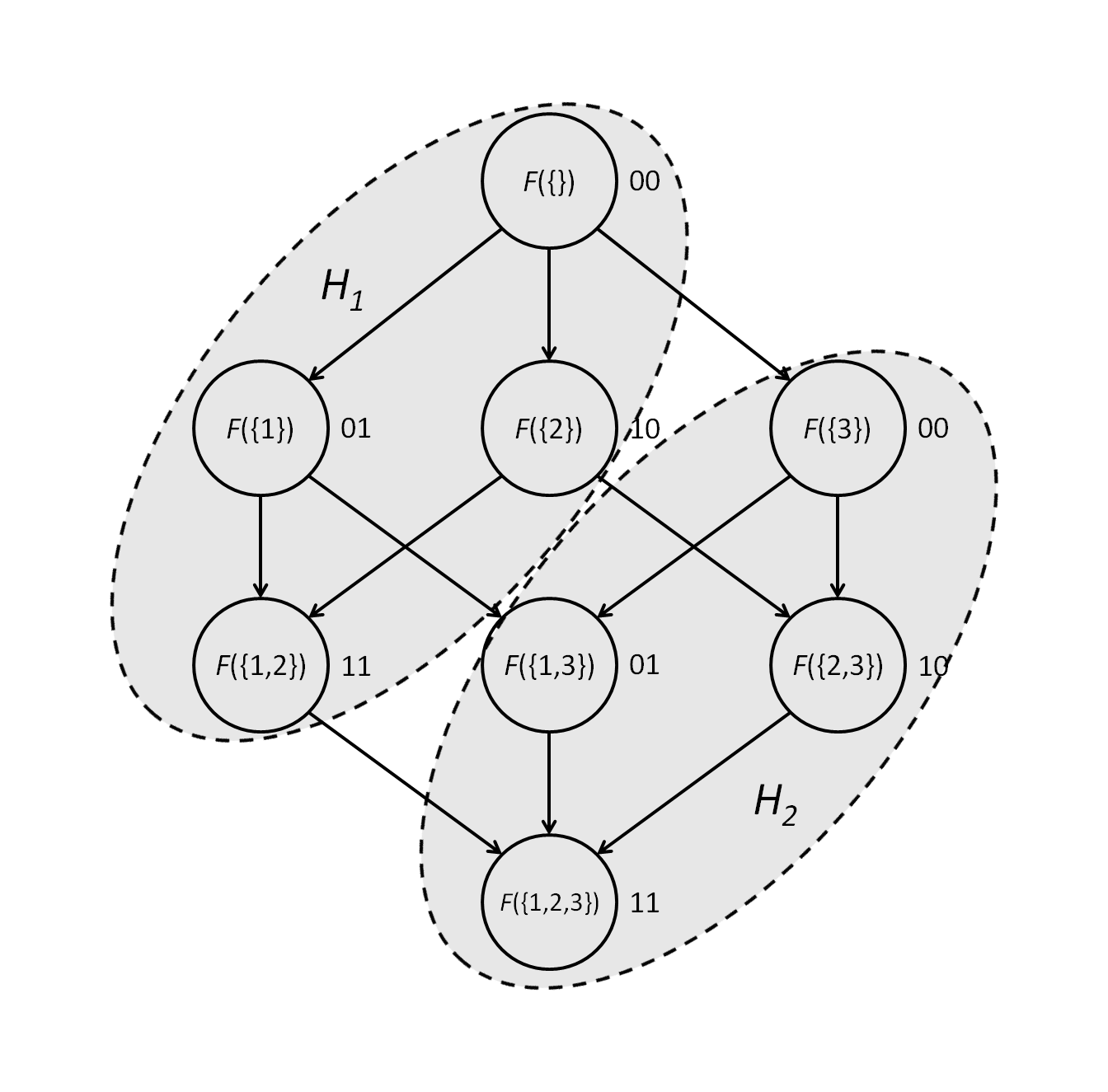}
		\caption{Computing $F(S)$}
		\label{fkdcube}
	\end{subfigure}
	\begin{subfigure}[b]{0.49\linewidth}
		\centering
		\includegraphics[width=1.0\linewidth]{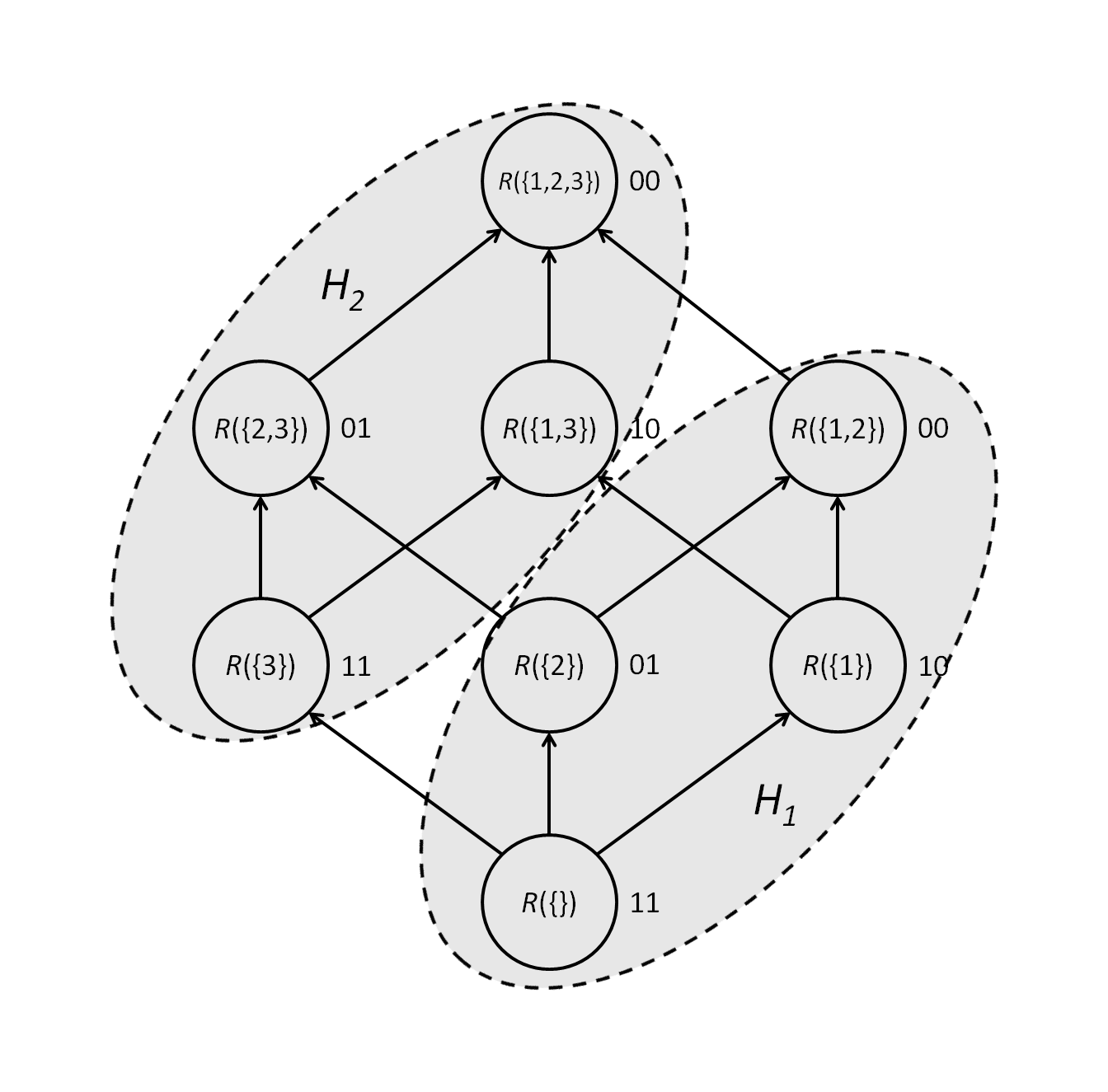}
		\caption{Computing $R(S)$}
		\label{rkdcube}
	\end{subfigure}			
	\caption{\small Pipelining execution of hypercubes to compute $F(S)$ and $R(S)$. The example shows a case with $n=3$ and $k=2$.}
	\label{pipecubes}
\end{figure}

Figure~\ref{fkdcube} illustrates a case of computing $F(S)$ with $n=3$ and $k=2$. In this example, the $3$-$D$ $F$ lattice is partitioned to two $2$-$D$ hypercubes $H_1$ and $H_2$. $H_1$ is processed before $H_2$ is processed.
One feature of the pipelining is that once a processor completes its computation in one $k$-$D$ hypercube, it transits to next $k$-$D$ hypercube immediately
without waiting for other processors to complete their computations in current hypercube. In Figure~\ref{fkdcube}, for example, once the processor $00$ completes node $\{\}$ and sends out data, 
it starts on node $\{3\}$ even if processors $01$, $10$, $01$ are still working on their nodes in $H_1$.
This feature prevents processors from excessive idling during the transitions between consecutive hypercubes.

The strategy to compute function $R(S)$ is similar. The only difference is the mapping of the subsets to processors. $R(S)$ is assigned to the processor with \emph{id} $r=\neg\omega[1,k]$ \footnote{$\neg\omega[1,k]$ denotes the bitwise complement of binary string $\omega[1,k]$.}, i.e., $R(S)$ is computed on the processor where $F(V-S)$ is computed. 
In other words, processors operate in the reverse order as that when they compute $F(S)$. An example of computing a $3$-$D$ $R$ lattice on $2$-$D$ hypercube is shown in Figure~\ref{rkdcube}.

\subsubsection{Overall Algorithm: ParaREBEL} 

With the $k$-$D$ algorithms for the two transforms, $A_i$ (and $B_i$) and $\Gamma_v$ functions can be computed efficiently. 
As mentioned, each processor with \emph{id} $r$ is responsible for computing $2^{n-k}$ subsets $S$ such that $r=\omega[1,k]$.
Note that before computing $\Gamma_v$, we need compute $q_v(S)F(S)R(V-\{v\}-S)$, where $F(S)$ and $R(V-\{v\}-S)$ are not necessarily on the same processor in the $k$-$D$ hypercube.
Fortunately, with our partition strategy, $F(S)$ and $R(V-\{v\}-S)$ locate either on the same processor or on the neighboring processors in the $k$-$D$ hypercube. Specifically, when $v\leq k$, processor $r$ with $r[v]=0$ need retrieve $R(S)$ from its neighbor  $r'=r\oplus2^{v-1}$; 
when $v > k$, $F(S)$ and $R(V-\{v\}-S)$ are on the same processor thus no message passing is needed to compute $q_v(S)F(S)R(V-\{v\}-S)$.

Finally, to compute $P(u\rightarrow v,D)$ for any $u,v\in S, u\neq v$, each processor $r$ first adds up all local $B_v(G_v)\Gamma_v(G_v)$ scores with $\omega_{G_v}[1,k]=r$, then a \texttt{MPI\_Reduce} is launched on
the $k$-$D$ hypercube to obtain the sum. The posteriors $P(u\rightarrow v|D)$ are evaluated as $P(u\rightarrow v,D)/F(V)$ on the processor $r$ with $r[i]=1$ for all $i\in\{1,...,k\}$.

\begin{algorithm}\small
	\caption{ParaREBEL computes the posterior probabilities of all $n(n-1)$ edges with $p=2^k$ processors.}
	\label{pararebel}
	\begin{algorithmic}[1]
		\Assumption each subset $S\subseteq V$ is encoded by an $n$-bit string $\omega$, where $\omega[i]=1$ if variable $i\in S$ and $\omega[i]=0$ otherwise. Each processor in the $k$-$D$ hypercube is encoded by an $k$-bit string $r$. 
		\State \textbf{for} each $i\in V$, compute $B_i(S)$ and $A_i(S)$ for all $S\subseteq V-\{i\}$ by \textbf{Algorithm~\ref{algorithm4}}. Each processor $r$ computes subsets $S$ s.t. $r=\omega[1,k]$.
		\State Compute $F(S)$ for all $S\subseteq V$ on $k$-$D$ hypercube. Each processor $r$ computes subsets $S$ s.t. $r=\omega[1,k]$.
		\State Compute $R(S)$ for all $S\subseteq V$ on $k$-$D$ hypercube. Each processor $r$ computes subsets $S$ s.t. $r=\neg\omega[1,k]$.
		\For {each $v\in V$}
		\If {$v\leq k$}
		\State Each processor $r$ with $r[v]=1$ sends all its $R$ scores to its neighbor $r'=r\oplus2^{v-1}$.
		\EndIf
		\State Each processor $r$ with $r[v]=0$ computes $q_v(S)F(S)R(V-\{v\}-S)$ for all its $S$. 
		\State Compute $\Gamma_v(G_v)$ for all $G_v\subseteq V-\{v\}$ with $|G_v|\leq d$ by \textbf{Algorithm~\ref{algorithm5}}.
		\For {each $u\in V-\{v\}$}
		\State Each processor $r$ recomputes $B_v(G_v)$ for all $G_v$ with $r=\omega_{G_v}[1,k]$,
		\NoNumber{then adds up all local $B_v(G_v)\Gamma_v(G_v)$ scores with $|G_v|\leq d$.}
		\State \texttt{MPI\_Reduce} is executed on the $k$-$D$ hypercube to compute the sum of all
		\NoNumber{$B_v(G_v)\Gamma_v(G_v)$, $P(u\rightarrow v,D)$, obtained on processor $r$ with $r[i]=1$ for all $i\in\{1,...,k\}$.} 
		\State Processor $r$ with $r[i]=1$ for all $i\in\{1,...,k\}$ evaluates $P(u\rightarrow v|D)=P(u\rightarrow v,D)/F(V)$.
		\EndFor 
		\EndFor
	\end{algorithmic}
\end{algorithm}

The overall $k$-$D$ hypercube algorithm, named as ParaREBEL\footnote{The serial algorithm in \citep{Koivisto06} is called REBEL.} (\emph{Parallel Rapid Exact Bayesian Edge Learning}), is outlined in Algorithm~\ref{pararebel}.

\subsubsection{Time and Space Complexity} 

We characterize the running time of \emph{ParaREBEL} under the assumption that the maximum in-degree $d$ is a constant.

For any fixed $i\in V$, computing $A_i(L_i)$ for all $L_i\subseteq V-\{i\}$ takes $O((d+1)\cdot 2^{n-k}+k(n-k)^d)$ time (Theorem~\ref{kdupzetatheory}). 
Thus, line 1 takes $|V|\cdot O((d+1)\cdot 2^{n-k}+k(n-k)^d)=O((d+1)n2^{n-k}+kn(n-k)^d)$ time to compute $B_i$ and $A_i$ scores for all $i\in V$. 

Line 2 and line 3 take $O(n(2^{n-k}+k))$ time each as we pipeline the execution of the $k$-$D$ hypercubes in $2^{n-k}+k$ steps and each step costs $O(n)$. 

In line 9, for any $v\in V$, computing $\Gamma_v$ scores takes $O((4d+4)\cdot 2^{n-k})$ time (Theorem~\ref{kddownzetatheory}).
Line 11 takes $O(\frac{n^d}{2^k})$ time as there are no more than $O(\frac{n^d}{2^k})$ $B_v(S)\Gamma_v(S)$ scores on each processor if bounded in-degree $d$ is assumed.
In line 12, \texttt{MPI\_Reduce} procedure takes
$O((\tau+\mu m)k)$ time. 
Thus, the time combined for Lines 4-15 is $O((((\tau+\mu m)k+\frac{n^d}{2^k})n+(4d+4)2^{n-k})\cdot n) = O(kn^2+\frac{n^{d+2}}{2^k}+4(d+1)n2^{n-k})=O(kn^2+4(d+1)n2^{n-k})$.\footnote{$O(kn^2+\frac{n^{d+2}}{2^k}+n2^{n-k})=O(kn^2+n2^{n-k})$ because $\frac{n^{d+2}}{2^k}$ is dominated by $n2^{n-k}$.} 

The total time for the overall algorithm is therefore $O(5(d+1)n2^{n-k}+kn(n-k)^d+kn^2)=O(5(d+1)n2^{n-k}+kn(n-k)^d)$.\footnote{We normally have $d\ge 2$, i.e., the up-bound of the in-degree is at least 2. In this case, $kn(n-k)^d$ dominates $kn^2$.}. 

Furthermore, $B$, $A$, $\Gamma$, $F$, $R$ scores are evenly distributed on the $2^k$ processors. Therefore, the storage per processor used by the parallel algorithm is $O(n2^{n-k})$. Since the space requirement of the sequential algorithm is $O(n2^n)$, our parallel algorithm achieves the optimal space efficiency.

In summary, we obtain the following results.
\begin{theorem}
\label{maintheorem}
Algorithm ParaREBEL runs in time $O(5(d+1)n2^{n-k}+kn(n-k)^d)$ and space $O(n2^{n-k})$ per processor. 
\end{theorem}

\section{Experiments}
In this section, we present the experiments for evaluating our ParaREBEL algorithm.
\subsection{Implementation and Computing Environment}
We implemented the proposed ParaREBEL algorithm\footnote{ParaREBEL is available for download at \url{http://www.cs.iastate.edu/~yetianc/software.html}.} in C++ and MPI and demonstrated its scalability on TACC Stampede\footnote{\url{http://www.tacc.utexas.edu/resources/hpc/stampede}}, a Dell PowerEdge C8220 cluster. Each computing node in the cluster consists of two Xeon Intel 8-Core E5-2680 processors (16 cores in all), sharing 32 GB memory. All experiments were run with one MPI process per core. To allow more memory per process, only 8 cores in each node were recruited so that each process could use up to 4 GB memory. The maximum number of nodes/cores allowed for a regular user on TACC Stampede is 256/4096.
To maintain 4 GB per core, we can only use up to 2048 cores. Thus, all the following experiments were done on up to 2048 cores.

\subsection{Running Time and Memory Usage}

We first evaluated the time and space complexity of our algorithm. We compared our implementation with REBEL\footnote{\url{http://www.cs.helsinki.fi/u/mkhkoivi/REBEL}}, a C++ implementation of the serial algorithm (Algorithm~\ref{algorithm1}) in \citep{Koivisto06}.

We generated a set of synthetic data sets with discrete random variables. Each dataset contains 500 samples. For each data set, we ran the serial algorithm and our ParaREBEL algorithm to compute the posterior probabilities for all $n(n-1)$ potential edges. We did two tests: one with varying bounded in-degree $d$ and fixed number of variables $n$, the other with varying number of variables $n$ and fixed bounded in-degree $d$. In both tests, the total running times were recorded and \emph{speedup} and \emph{efficiency} were computed.
In the second test, the memory usages per processor were collected and the total memory usages were calculated.

In the first test, we fixed $n=25$ and studied the performance of ParaREBEL algorithm with respect to the bounded in-degree $d$ ($d=2,4,6,8$). The run-times are presented in Table~\ref{varydtable}. The corresponding speedups and efficiencies are illustrated in Figure~\ref{varydplot}. Generally, we observed overall good scaling (see speedup plot in Figure~\ref{varydplot}) for all values of $d$.
The speedup and efficiency both improve when $d$ increases from 2 to 4, but decline when $d$ keeps increasing from 4 through 6 to 8.
From our theoretical analysis of running time , we have $speedup=\frac{2(d+1)n2^n}{5(d+1)n2^{n-k}+kn(n-k)^d}=\frac{2\cdot2^n}{5\cdot2^{n-k}+\frac{k(n-k)^d}{d+1}}$ and \emph{efficiency}$=\frac{2(d+1)n2^n}{5(d+1)n2^n+kn(n-k)^d2^k}=\frac{2\cdot2^n}{5\cdot2^n+\frac{k(n-k)^d}{d+1}2^k}$.
Both formulas are not a monotonic function of $d$.
when $d$ is small, $d+1$ in the denominator of $\frac{k(n-k)^d}{d+1}$ dominates thus both speedup and efficiency improve when $d$ increases. 
When $d$ is large, $k(n-k)^d$ starts to dominate and the two measures decline when $d$ increases.
Thus, our empirical result is consistent with our theoretical result. 
For $d=4$, the efficiencies are maintained above 0.53 with up to 2048 cores.\footnote{Generally, parallel algorithms with \emph{efficiency}$\ge 0.5$ are considered to be successfully parallelized.}

In the second test, we fixed $d=4$ and studied the performance of the algorithm with respect to the number of variables ($n=21, 23, 25, 27, 29, 31, 33$).
We first compared the run-times. As showed in Table~\ref{varyntable}, the run-times are reflective of the exponential dependence on $n$. 
Further, we observed that the algorithm scaled much better when $n$ becomes larger (see speedup and efficiency plot in Figure~\ref{varynplot}). 
This is also supported by our theoretical result.
With a minor transform, our running time analysis suggests $speedup=\frac{2(d+1)}{5(d+1)2^{-k}+k(n-k)^d2^{-n}}$.
When $n$ is large enough, speedup (and efficiency) is a increasing function of $n$.
For $n=25$, the parallel algorithm maintains an efficiency of about 0.6 with up to 2048 cores. 
For $n=33$, the problem can only be solved on 1024 and 2048 cores due to memory constraint. 
We had a try on $n=34$ using 2048 cores but were not able to solve it as it ran out of memory.

\begin{table}[h] \small
\caption{Run-time for the test data with $n=25$ with varying bounded in-degree $d$.}
\label{varydtable} 
\begin{center} 
\begin{tabular}{lllll}
\hline
\multicolumn{1}{c}{\bf No.CPUs} &\multicolumn{4}{c}{\bf Run-time (seconds)} \\
 &$d=2$ &$d=4$ &$d=6$ &$d=8$\\
\hline
Serial &1319	&2295	&4308	&7739 \\
4	&1284		&1330	&1500	&2383 \\
8	&575		&594	&711	&1304 \\
16	&327		&338	&417	&764 \\
32	&139		&146	&181	&466 \\
64	&59.9		&64.2	&102	&268 \\
128	&26.6		&29.4	&55.6	&153 \\
256	&11.7		&13.8	&31.3	&86.8 \\
512	&5.2		&6.8	&18.2	&48.5 \\
1024	&2.5	&3.6	&11.0	&26.9 \\
2048	&1.5	&2.1	&6.7	&14.8 \\
\hline
\end{tabular} 
\end{center} 
\end{table}

\begin{figure}[!htb]
\centering
\includegraphics[width=75mm]{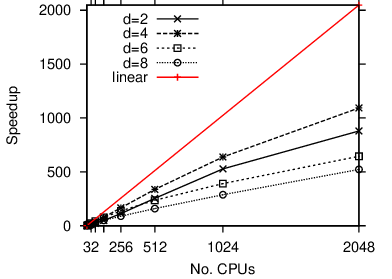}
\includegraphics[width=75mm]{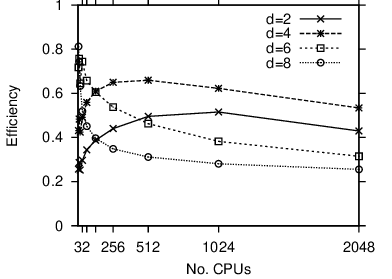}
\caption{\emph{Speedup} and \emph{efficiency} for the test data set with $n=25$ with varying bounded in-degree $d$. The red diagonal line in speedup plot represents the linear or ideal speedup, i.e., the up-bound that a parallel algorithm can achieve in theory.}
\label{varydplot}
\end{figure}

One interesting observation is that for any fixed $d$ and $n$, the parallel efficiency increases as the No.CPUs increases, peaks at somewhere in between, then gradually decreases as No.CPUs goes
up to 2048 CPUs (see efficiency plot in Figure~\ref{varynplot}). Mathematically, this optimum can be found by maximizing \emph{efficiency}$=\frac{2(d+1)2^n}{5(d+1)2^n+k(n-k)^d2^k}$, i.e., minimizing $k(n-k)^d2^k$ over $k$. 
Solving this optimization problem yields $k^*=n(\ln2+1)/(\ln2+1+d)\approx 1.7n/(d+1.7)$. Plugging in $n=25$ and $d=4$ yields $k^*=7.5\approx 8$, i.e., $2^{k^*}\approx256$ cores.
Plugging in $n=23$ and $d=4$ yields $k^*=6.8\approx 7$, i.e., $2^{k^*}\approx128$ cores. All these results consist exactly with the observation in Figure~\ref{varynplot}.
This provides another piece of solid experimental evidence for Theorem~\ref{maintheorem}. Further, this optimum $k^*$ is proportional to $n$, i.e., the optimal \emph{efficiency} will be achieved by using larger number
of cores when problem becomes larger. This suggests our ParaREBEL algorithm scales very well with respect to the problem size $n$. 

\begin{table}[h]  \small
\caption{Run-time for the test data sets with $n=21, 23, 25, 27, 29, 31, 33$ with fixed $d=4$.}
\label{varyntable} 
\begin{center} 
\begin{tabular}{llllllll}
\hline
\multicolumn{1}{c}{\bf No.CPUs} &\multicolumn{7}{c}{\bf Run-time (seconds)} \\
 &$n=21$ &$n=23$ &$n=25$ &$n=27$ &$n=29$ &$n=31$ &$n=33$\\
\hline
Serial &96.5	&492 &2295 &{-} &{-} &{-} &{-}\\
4 &44.1	&252 &1330 &{-} &{-} &{-}  &{-} \\
8 &17.2	 &94.2 &594 &{-} &{-} &{-} &{-}\\
16 &10.3   &55.5 &338  &{-} &{-} &{-} &{-}\\
32 &5.0	&25.5 &146  &682 &{-} &{-} &{-}\\
64 &2.7	&11.9 &64.2  &385 &2201 &{-} &{-}\\
128 &1.6	&5.8 &29.4 &167 &864 &{-}\\
256 &0.97	&3.2	&13.8	&73.5 &389	&2540 &{-} \\
512 &0.61	&1.8	&6.8	&33.9 &196	&987 &{-} \\
1024 &0.4	&1.1	&3.6	&15.9 &87	&488 &2884 \\
2048 &0.27	&0.7	&2.1	&7.8 &39	&215 &1452 \\
\hline
\end{tabular} 
\end{center} 
\end{table} 

\begin{figure}[!htb]
\centering
\includegraphics[width=75mm]{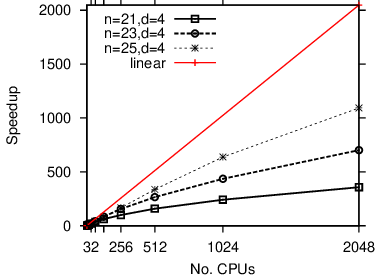}
\includegraphics[width=75mm]{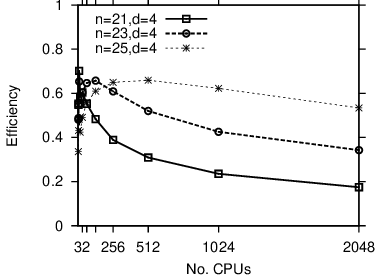}
\caption{\emph{Speedup} and \emph{efficiency} for the test data sets with $n=21, 23, 25$. The red diagonal line in speedup plot represents the linear or ideal speedup, i.e., the up-bound that a parallel algorithm can achieve in theory.}
\label{varynplot}
\end{figure}

\begin{table}[h]
\small
\caption{Memory usage for the test data with $n=23, 25, 27, 29, 31, 33$ with fixed $d=4$. The term outside the parentheses is the total memory usage measured in GB, the term inside the parentheses is the memory usage per core measured in MB. The missing entries indicate the cases where the program runs out of memory.}
\label{memorytable} 
\begin{center} 
\begin{tabular}{lllllll}
\hline
\multicolumn{1}{c}{\bf No.CPUs} &\multicolumn{6}{c}{\bf Memory Usage} \\
&$n=23$ &$n=25$ &$n=27$ &$n=29$ &$n=31$ &$n=33$\\
\hline
4 &1.88 (481) &8.00 (2049) &{-} &{-} &{-}  &{-} \\
8 &1.88 (240) &8.00 (1025) &{-} &{-} &{-} &{-}\\
16 &1.88 (121) &8.01 (513) &{-} &{-} &{-} &{-}\\
32 &2.17 (70) &8.30 (266)  &34.32 (1098) &{-} &{-} &{-}\\
64 &2.49 (40) &8.62 (138)  &34.64 (554) &144.68 (2315) &{-} &{-}\\
128 &3.31 (27) &9.46 (76) &35.48 (284) &145.53 (1164) &{-} &{-}\\
256 &4.93 (20)	&11.06 (44)	&37.09 (148) &147.07 (588)	&606.13 (2425) &{-} \\
512 &8.40 (17)	&14.73 (30)	&40.76 (82) &150.87 (302)	&615.03 (1230) &{-} \\
1024 &17.58 (18)	&23.62 (24)	&49.72 (50) &159.87 (160)	&623.88 (624) &2520 (2520) \\
2048 &41.08 (21) 	&47.32 (24)	&72.97 (36) &183.69 (92)	&647.27 (324) &2560 (1300) \\
\hline
\end{tabular} 
\end{center} 
\end{table}

We then examined the actual memory usages with respect to the number of variables $n$ and the number of cores $2^k$ in Table~\ref{memorytable}.
For $n=23$, the total memory usage remains the same (1.88 GB) for $2^k=4, 8, 16$ cores, but starts to increase as the number of cores increases from 16 to 2048.
This increase is dramatic for the number of cores ranging from 256 to 2048, i.e, the memory usage is doubled when the number of cores is doubled.
This can be explained by examining the memory usage per core. For $2^k=4, 8, 16$, the memory usage per core decreases by half when the number of cores is doubled.
This is consistent with our theoretical analysis that the space complexity is $O(n2^{n-k})$ per core. When $2^k\ge 16$, the reduction slows down and the memory usage plateaus at about $20$ MB per core.
It is speculated that in addition to the memory allocated for storing the $B, A, F, R, \Gamma$ scores, each core requires extra $10\sim20$ MB memory to store program execution related data
in order to run the program. This overhead is negligible when the memory usage per core is dominated by the scores but comes into play otherwise.
For $n=25$, total memory usage stays at about 8 GB for $2^k=4 \sim 64$ and starts to increase thereafter; 
for $n=27$, total memory usage stays at about 35 GB for $2^k=32\sim 256$ and starts to increase thereafter;
for $n=29,31,33$, the memory usage per core is dominated by the scores, thus, the total memory usage stays roughly constant with respect to the number of cores examined.
Further, it is easily observed that the memory usages (total memory usage and memory usage per core) are reflective of the exponential dependence on $n$.
Thus, the observations on the memory usage are consistent with our analysis of the space complexity. 

Moreover, the missing entries in the table are the cases where the program runs out of memory. Thus, we concluded that it requires at least 4 GB memory per core if $n-k > 23$. To solve a problem of $n\ge 34$, we need 2048 cores with more than 4 GB memory per core or 4096 cores with more than 2 GB memory per core. 
However, these resources are unavailable to a regular user on TACC Stampede. 
Further, we observed that the problem of $n=33$ could be solved on 1024 cores in less than one hour, and 2048 cores in less than half an hour. 
The computing times are still far away from the practical limit.
Thus, memory requirement is still the bottleneck that determines the feasibility limit in practice.

\subsection{Knowledge Discovery}

Finally, we applied our algorithm to a biological dataset for discovering the regulatory network responsible for controlling the expression of various genes involved in {\it Saccharomyces cerevisiae} (\emph{yeast}) pheromone response pathways \citep{hartemink2001principled}.
This data set consists of 33 variables, of which 32 variables represent discretized levels of gene expression 
and an additional binary variable represents the mating type of various haploid strains of yeast.
A total number of 320 observations are recorded.
Bayesian network structure models for this data set have been constructed by using model selection methods such as greedy hill climbing, simulated annealing or by Bayesian model averaging over models selected during the simulated annealing \citep{hartemink2001principled,hartemink2002combining}.

\begin{figure}
	\centering
	\includegraphics[width=0.9\linewidth]{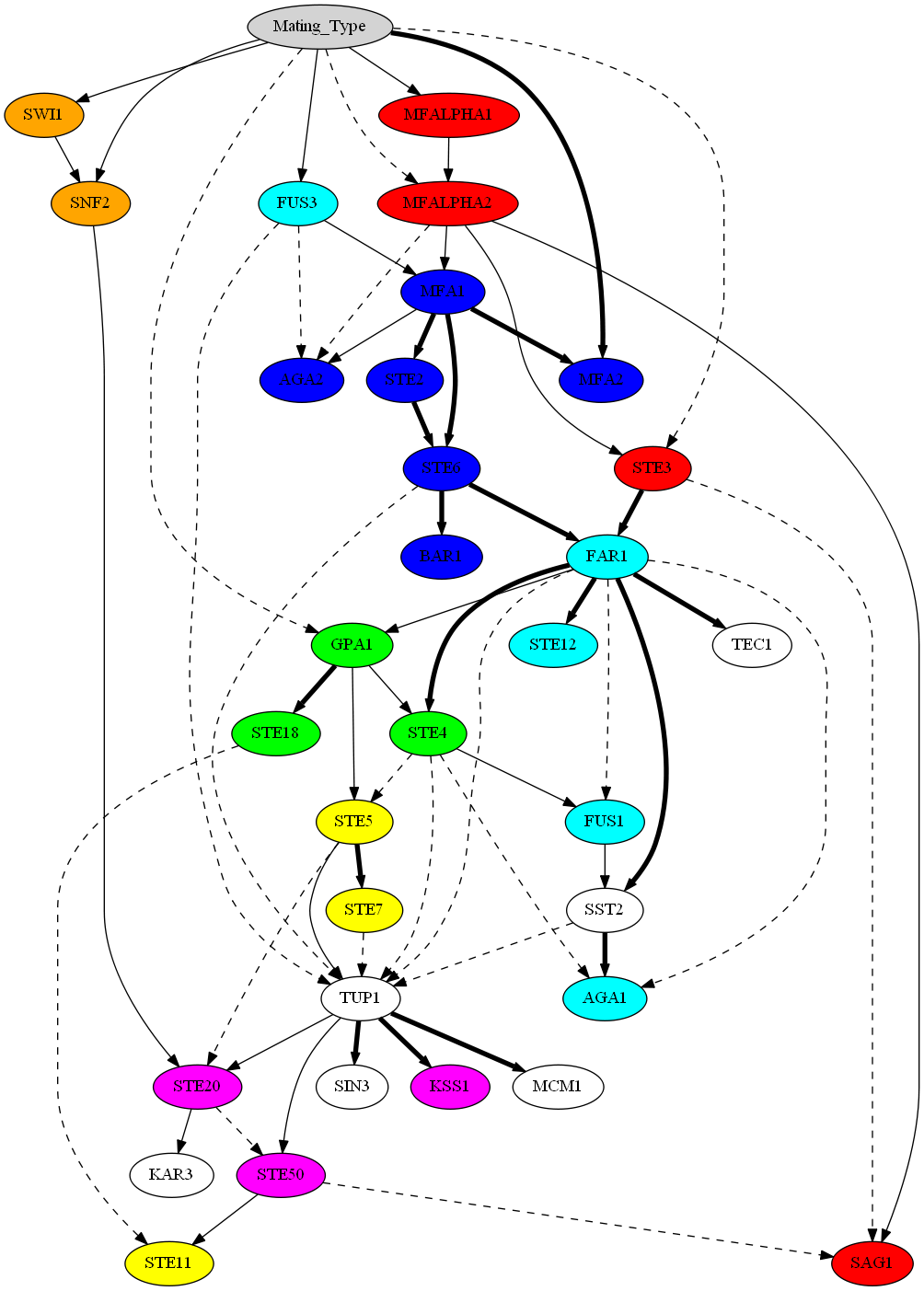}
	\caption{\small Network model learned for the \emph{yeast} pheromone response pathways data set.
		Nodes have been augmented with color information to indicate the different groups of variables with known relationships in the literature.
		Directed edges are formatted according to their posterior probabilities: heavily weighted ($posterior\ge 0.9$), solid ($0.5\le posterior<0.9$), and dashed ($0.1\le posterior<0.5$).}
	\label{yeast}
\end{figure}

We used our ParaREBEL algorithm to compute the exact posterior probabilities of all 1056 potential edges.
The total running time was 1542 seconds on 2048 cores.
We then constructed a network that consisted of (important) edges whose posteriors were greater than 0.1 (we set this threshold such that the constructed network is a DAG).
The network model consists of 60 edges and is illustrated in Figure~\ref{yeast}.
Nodes have been augmented with color information to indicate the different groups of variables with known relationships in the literature.
Edges are formatted according to their posterior probabilities. 

Since the ground truth network is unknown, we cannot evaluate the accuracy of the model.
However, we observe a number of interesting properties.
First, variables in the same group (with the same color) tend to form a cluster (directly connected subgraph) in the network and the intra-class edges are generally more probable than the inter-class edges. This demonstrates that our algorithm is capable of recovering the (important) interactions in the \emph{yeast} pheromone response pathways.
Second, the \texttt{Mating\_Type} variable is at the source of the network,
and contributes to the ability to predict the state of a large number of variables, which is to be expected.
Further, in \citep{hartemink2001principled}, two types of models were learned, one obtained using greedy or simulated annealing search without any domain constraint (see Figure 7-3 in \citep{hartemink2001principled}), the other learned using the similar search approaches but with constraints governing the inclusion and exclusion of edges which were derived from genomic analysis (see Figure 7-4 in \citep{hartemink2001principled}). Interestingly, our network, which was constructed without any domain constraints, is more like the model learned with the constraints. 
This suggests that the network constructed with edge posteriors may achieve better modeling of the regulatory network than the model learned using model selection methods. Future research could explore additional data sets to confirm this observation.

\section{Discussions and Conclusions}
Exact Bayesian structure discovery in Bayesian networks requires exponential time and space. 
In this work, we have presented a parallel algorithm capable of computing the exact posterior probabilities for all $n(n-1)$ potential edges
with optimal time and space efficiency.
To our knowledge, this is the first practical parallel algorithm for computing the exact posterior probabilities of structural features in BNs. 
We demonstrated its capability on datasets with up to 33 variables and its scalability on up to 2048 processors. 
To our knowledge, 33-variable network is the largest problem solved so far.
We have also applied our algorithm to a biological data set for discovering the (\emph{yeast}) pheromone response pathways. This demonstrated our algorithm in the task of knowledge discovery.

Our algorithm makes twofold algorithmic contributions. First, it achieves an efficient parallelization of the base serial algorithm by presenting a delicate way to coordinate the computations of correlated DP procedures
such that large amount of data exchange is suppressed during the transitions between these DP procedures. Second, it develops two parallel techniques
for computing two variants of well-known \emph{zeta transform}. These features or ideas can potentially be extended and applied in developing parallel algorithms for related problems.
For example, the algorithm in \citep{tian2009computing} involves similar steps and transforms.
Further, as zeta transforms are fundamental objects in combinatorics and algorithmics, 
the parallel techniques developed here would also benefit the researches beyond the context of Bayesian networks \citep{bjorklund2007fourier,bjorklund2010trimmed,nederlof2009fast}.

From the experiments, we observed that memory requirement reached the limit much faster than computing time did. 
Thus, one of the future work is to improve the algorithm such that less space is used.
Particularly, there is a possibility to combine the present algorithm with the method in \citep{parviainen2010bayesian} to trade space against time.
 \\
 \\
{\noindent ParaREBEL is available at \url{http://www.cs.iastate.edu/~yetianc/software.html}.}

\acks{
This work used the Extreme Science and Engineering Discovery Environment (XSEDE), which is supported by National Science Foundation grant number ACI-1053575.
}




\bibliography{BN}

\end{document}